\documentclass[journal=jacsat,manuscript=article]{achemso}
\usepackage{amsmath}
\usepackage{amssymb}
\usepackage[version=3]{mhchem} 

\title{Brain-Inspired Reservoir Computing Using Memristors with Tunable Dynamics and Short-Term Plasticity}

\author{Nicholas X. Armendarez}
\affiliation[Pennsylvania State University]{Department of Mechanical Engineering, The Pennsylvania State University, 336 Reber Building, University Park, PA 16802}
\author{Ahmed S. Mohamed}
\affiliation[Pennsylvania State University]{Department of Mechanical Engineering, The Pennsylvania State University, 336 Reber Building, University Park, PA 16802}
\author{Anurag Dhungel}
\affiliation{Department of Electrical and Computer Engineering, The University of Mississippi, 310 Anderson Hall, University, MS 38677}
\author{Md Razuan Hossain}
\affiliation{Department of Electrical and Computer Engineering, The University of Mississippi, 310 Anderson Hall, University, MS 38677}
\author{Md Sakib Hasan\textsuperscript{*}}
\affiliation{Department of Electrical and Computer Engineering, The University of Mississippi, 310 Anderson Hall, University, MS 38677}
\author{Joseph S. Najem}
\affiliation[Pennsylvania State University]{Department of Mechanical Engineering, The Pennsylvania State University, 336 Reber Building, University Park, PA 16802}
\email{mhasan5@olemiss.edu, jsn5211@psu.edu}

\begin{document}

\begin{abstract}

Recent advancements in reservoir computing research have created a demand for analog devices with dynamics that can facilitate the physical implementation of reservoirs, promising faster information processing while consuming less energy and occupying a smaller area footprint. Studies have demonstrated that dynamic memristors, with nonlinear and short-term memory dynamics, are excellent candidates as information-processing devices or reservoirs for temporal classification and prediction tasks. Previous implementations relied on nominally identical memristors that applied the same nonlinear transformation to the input data, which is not enough to achieve a rich state space. To address this limitation, researchers either diversified the data encoding across multiple memristors or harnessed the stochastic device-to-device variability among the memristors. However, this approach requires additional pre-processing steps and leads to synchronization issues. Instead, it is preferable to encode the data once and pass it through a reservoir layer consisting of memristors with distinct dynamics. Here, we demonstrate that ion-channel-based memristors with voltage-dependent dynamics can be controllably and predictively tuned through voltage or adjustment of the ion channel concentration to exhibit diverse dynamic properties. We show, through experiments and simulations, that reservoir layers constructed with a small number of distinct memristors exhibit significantly higher predictive and classification accuracies with a single data encoding. We found that for a second-order nonlinear dynamical system prediction task, the varied memristor reservoir experimentally achieved an impressive normalized mean square error of $1.5\times 10^{-3}$, using only five distinct memristors. Moreover, in a neural activity classification task, a reservoir of just three distinct memristors experimentally attained an accuracy of 96.5 \%. This work lays the foundation for next-generation physical reservoir computing systems that can exploit the complex dynamics of its diverse building blocks to achieve increased signal processing capabilities. 

\end{abstract}

\section{Introduction}
\begin{figure}
  \includegraphics[width=\linewidth]{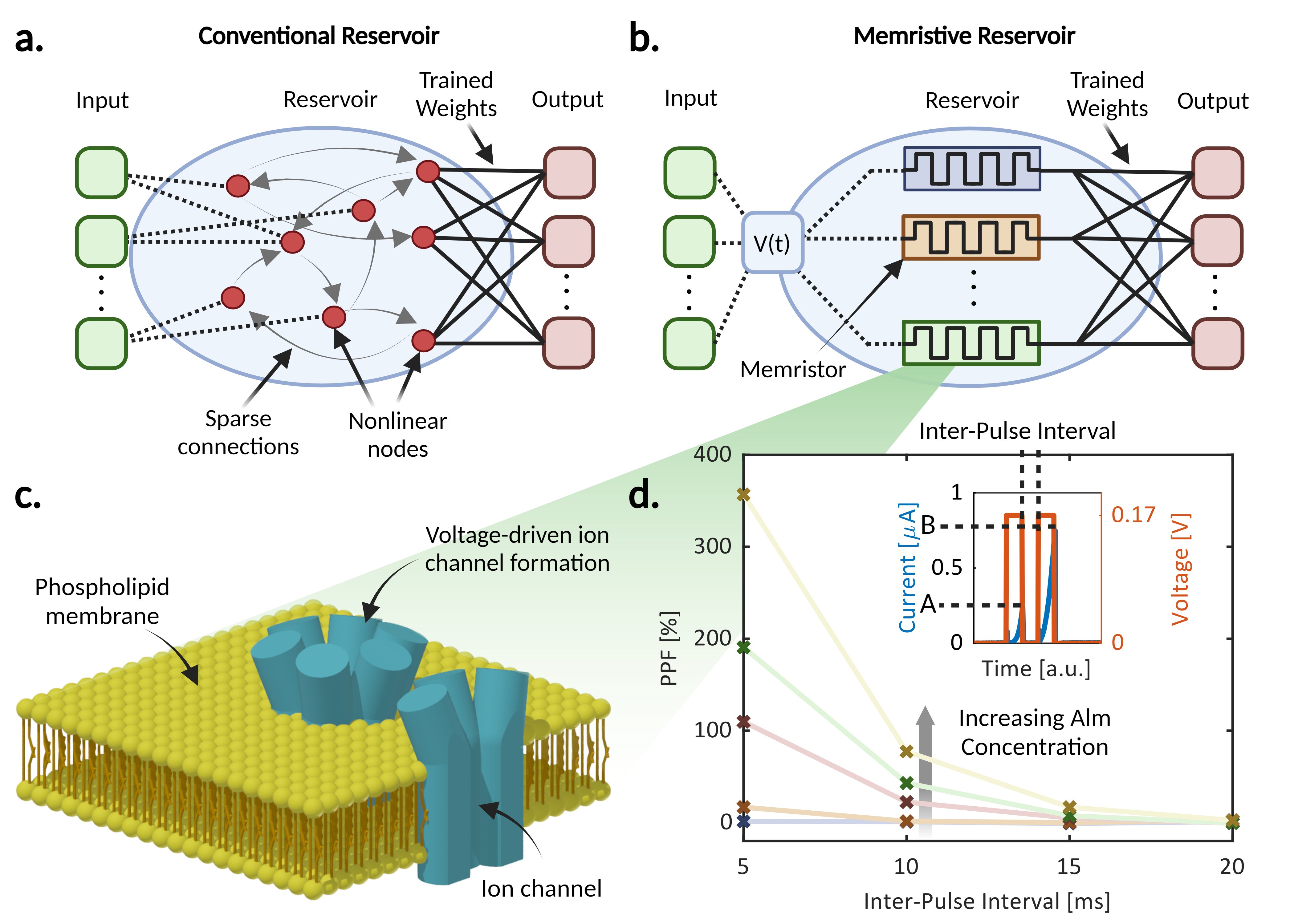}
  \caption{(a) The schematic depicts a conventional reservoir computer comprising an input layer, a recurrent neural network-based reservoir layer, and an output layer. (b) An equivalent system can be constructed using diverse dynamic memristors. In this system, data is initially encoded into a voltage waveform $V(t)$ that can be fed to the memristors. At regular intervals, the conductance states of the memristors are sampled to determine the reservoir states. We use diverse nonlinear conductance responses from multiple devices to achieve high-dimensional projection, incorporating recurrence through the memristors' volatile memory. (c) A schematic illustrating the structure of our ion-channel-based memristor where alamethicin peptides form pores within an insulating lipid membrane, altering its conductance. Below a voltage threshold that is dependent on peptide concentration, these peptides horizontally adhere to the outer surface of the membrane. However, an applied voltage above the threshold compels them into a vertically inserted position, as depicted. Once inserted, they establish conductive ion channels, thereby enhancing the overall membrane conductivity. (d) We demonstrate PPF for five concentrations of alamethicin, ranging from 1 $\mu$M to 3 $\mu$M in 0.5 $\mu$M increments. The inset displays a typical memristor current response to two square voltage pulses administered in close temporal proximity, separated by an inter-pulse interval (IPI). The pulses generate an amplified current response, illustrating the PPF of the memristors. Quantitatively, PPF is defined in terms of the peaks of the first pulse (\textit{A}) and second pulse (\textit{B}) as follows: $PPF=(B-A)/A\times100\%$. The PPF values for the five alamethicin concentrations are then plotted against IPI. The voltage pulses employed were 5 ms in duration and square waveforms with an amplitude of 170 mV.}
  \label{fig:fig1}
\end{figure}

Reservoir Computing (RC) is an emerging neural network framework that excels in time series prediction and temporal data classification \cite{jaeger2001echo, jaeger2002adaptive, maass2002real} by extracting features from an input signal and mapping them into a higher dimensional state space \cite{jaeger2001echo, jaeger2002adaptive, maass2002real}. The resulting states from the reservoir layer can then be fed into a trained readout layer to predict or classify the outputs of interest \cite{jaeger2001echo,jaeger2002adaptive}. To date, the reservoir layers of this highly efficient computing architecture have been realized using various physical substrates, such as conventional circuit components \cite{appeltant2011information, liang2022rotating} or photonic elements \cite{nakajima2021scalable}. Recently, studies have demonstrated that dynamic memristive devices can satisfy the requirements for reservoir computing while offering increased functional density and energy efficiency \cite{tanaka2019recent, du2017reservoir, midya2019reservoir, moon2019temporal}. Dynamic memristors are nonlinear, two-terminal electrical components that exhibit short-term plasticity or memory (i.e., modulation of response to repeated stimuli)\cite{zucker1989short} and can change their conductance based on both present and past activities \cite{chua1971memristor}. However, when the input stimulus is removed, these memristors gradually restore their resting conductance state \cite{wang2020recent}. Memristors that have been employed as reservoir layers include perovskite halide \cite{zhu2020memristor, xiao2020recent}, biomolecular \cite{najem2018memristive, maraj2023sensory}, and metal-oxide memristors \cite{du2017reservoir, moon2019temporal, midya2019reservoir}, to name a few. Unlike conventional reservoirs that employ recurrent spatial nodes to map the fed inputs to a higher-dimensional feature space \cite{jaeger2002adaptive}, dynamic memristor-based reservoirs leverage the inherent short-term memory and nonlinear dynamics of the memristors. They temporally couple their internal states, often referred to as ``virtual nodes" \cite{moon2019temporal}, to nonlinearly transform input stimuli into a higher-dimensional feature space \cite{moon2019temporal}. To achieve effective higher-dimensional mapping, the resulting state responses are required to be linearly independent for maximum input separability \cite{cover1965geometrical}. In conventional reservoir computing (Figure \ref{fig:fig1}a), linearly-independent state responses can be achieved by engineering spatial nodes with nonlinear activation functions and by selecting a random, sparse, inter-node connection matrix \cite{jaeger2002adaptive,ozturk2007analysis}. However, most physical memristor-based reservoirs to date rely on input encoding techniques, such as pulse-width encoding and masking \cite{du2017reservoir, maraj2023sensory, zhong2021dynamic}, as well as stochastic device-to-device variations \cite{moon2019temporal}, to achieve a high-dimensional, linearly-independent state space extracted from the temporal virtual nodes. This requirement stems from the use of nominally identical memristors, which produce similar outputs in response to the same stimulus \cite{moon2019temporal}.

Although these approaches have demonstrated success for specific tasks \cite{du2017reservoir,zhong2021dynamic,moon2019temporal}, they suffer from some fundamental limitations. An ideal physical reservoir must exhibit nonlinearity and fading memory \cite{tanaka2019recent}. Ideally, each state within the reservoir should offer a unique nonlinear transformation at different timescales. Using different pulse-encoding techniques \cite{du2017reservoir} addresses the multiple timescale issue but lacks unique nonlinearity. While masking \cite{zhong2021dynamic} introduces unique nonlinearity during pre-processing, it incurs significant overhead and does not vary the timescale. Employing multiple memristors with identical intrinsic nonlinearity and fading memory alongside these pre-processing techniques results in very similar reservoir states. This approach, in turn, necessitates a larger reservoir with reduced accuracy and increased implementation overhead, affecting both training and inference. The excessive high dimensionality with strong correlation can also lead to overfitting, particularly with smaller training datasets. Furthermore, the pre-processing steps required for each memristor in a parallel reservoir can pose synchronization challenges for real-time tasks  \cite{zhu2020memristor}. Therefore, a more efficient approach entails encoding the input data at a single timescale without variable masking and sending the resulting signal to all memristors in the reservoir layer. These memristors must exhibit different dynamics to yield various responses to the same signal, thus achieving the same high-dimensional projection as other methods \cite{carbajal2015memristor}. Such an approach would reduce the need for extensive input encoding and enable synchronized conductance sampling rates. Interestingly, this concept finds biological inspiration in the diversity of synapses connecting biological neurons. In this context, a single pulse train of action potentials can be transmitted through multiple synaptic contacts with a variety of magnitudes and temporal profiles, depending on the speed of facilitation or depression of each synapse \cite{abbott2004synaptic, fortune2001short, george2011diversity}. According to a recent study, this phenomenon enhances the spatiotemporal pattern recognition capabilities of single neurons \cite{beniaguev2022multiple}. To achieve the same capabilities in artificial reservoir computing systems, it is crucial to use memristors with variable dynamic properties. This variability ensures that each memristor responds significantly differently to the same input, thereby achieving the desired high-dimensional projection. These memristors must be made from easily tunable materials that can be intentionally designed to exhibit different intrinsic timescales and dynamics. However, to the best of our knowledge, current dynamic memristors, particularly metal-oxide variants used for reservoir computing applications, have not yet been controllably designed to exhibit varied dynamic properties to create a diverse reservoir layer. Conversely, soft systems, such as biomolecular memristors \cite{najem2018memristive, maraj2023sensory}, offer both high biomimicry and tunability. Biomolecular, ionic, and nanofluidic memristors \cite{xiong2023neuromorphic, xie2022perspective, robin2023long} have recently gained attention due to their chemical similarity, operational mechanism, and operating voltages resembling natural synapses.

In this study, we use ion-channel-based memristors (Figure \ref{fig:fig1}c), which consist of phospholipid membranes doped with voltage-driven ion-channel-forming alamethicin peptides \cite{najem2018memristive}, to construct a diverse reservoir layer. We investigate how varying the concentration of alamethicin peptides allows us to tune multiple memristors to exhibit diverse dynamic time constants and nonlinearity. Furthermore, for memristors with the same alamethicin concentration, we demonstrate their unique property of voltage-dependent time constants, where their time constants are exponential functions of the present voltage applied. This functional dependence on voltage enables further device response tuning by applying a constant DC voltage offset to the signal. When combined with the base encoding data, this voltage offset produces distinct responses to the same relative input, both in terms of nonlinearity and fading memory. To verify the uniqueness of outputs for each memristor, we tested the percentage increase in conductance between consecutive pulses, a measure known as paired-pulse facilitation (PPF) (Figure \ref{fig:fig1}d). To experimentally validate the main hypothesis of this study, which posits that utilizing a reservoir with various intrinsic decay times enhances prediction and classification performance by introducing additional state-space diversity, we construct reservoir layers using memristors with distinct dynamics (Figure \ref{fig:fig1}b). Firstly, we solved a dynamic model prediction task for a second-order nonlinear dynamical system using a reservoir made of five parallel memristors, each with a unique concentration of alamethicin, resulting in varied responses. Secondly, we exploited the voltage-dependent time constants of the devices to classify neural activity patterns by manipulating the DC voltage offset to three parallel memristors. Our results demonstrate that this approach achieves improved prediction and classification accuracy compared to current state-of-the-art memristor reservoirs, while also requiring fewer trained weights, thus reducing the computational demands of supervised training. By employing a reservoir composed of controllably modular memristors, we eliminate the need for individualized extensive pre-processing and reduce the reliance on unpredictable device-to-device variations to obtain a reservoir exhibiting richer dynamics with a sufficiently large set of output states with unique fading memory and nonlinearity.

\section{Results}

\subsection{Characterizing the memory and nonlinear dynamics in alamethicin memristors}

To serve as reservoir computing elements, memristors must exhibit some nonlinear dynamics in addition to fading memory (i.e., short-term memory) \cite{du2017reservoir}. Here, we evaluate the memory and nonlinear properties of ion-channel-based memristive devices made with varying alamethicin concentrations at different voltage bias levels. The memristors used in this study consisted of artificial phospholipid membranes doped with alamethicin peptides (see the Methods Section for details), as previously described by Najem \textit{et al.} \cite{najem2018memristive, najem2019assembly}. At sufficiently low alamethicin concentrations and sub-threshold transmembrane potentials (varies according to membrane composition \cite{najem2018memristive}), these peptides adsorb parallel to the surface of the membrane without impacting its insulating conductance state. However, in the presence of supra-threshold transmembrane potentials, the alamethicin peptides insert into the membrane driven by their dipole moment \cite{okazaki2003ion}. Once inserted, a small number of peptides then agglomerate to form conductive ion channels \cite{eisenberg1973nature}. As a result, the overall conductance of the membrane increases and is directly proportional to the number of ion channels formed within the membrane \cite{eisenberg1973nature}. Upon the removal of the externally applied electric field, the peptides spontaneously exit the membrane and return to the surface, reverting the membrane to an insulating state. Previous studies have demonstrated that these channels exhibit transient hysteresis when subjected to dynamic voltage inputs, resulting from a mismatch between the rates of channel formation and decay \cite{najem2018memristive}. It is this pinched hysteresis in the I-V plane that classifies these devices as dynamic memristors according to Leon Chua's definition \cite{chua1971memristor}.

The dynamics of channel formation and decay are affected by a number of factors, including alamethicin concentration, applied voltage amplitude, salt concentration and species, solution pH, membrane type, and temperature \cite{najem2018memristive, okazaki2003ion, eisenberg1973nature, boheim1978analysis,  mcclintic2023heterosynaptic}. In this study, we specifically investigate the effects of voltage offset and alamethicin concentration on the transient dynamics of the memristors, as these factors have the most significant impact on the device behavior. While the aforementioned factors may enhance the tunability of the alamethicin memristors, each presents inherent limitations. Varying salt concentration may yield similar effects to varying alamethicin concentration, but can also destabilize the lipid membrane at very low concentrations, restricting the achievable range of tunable states \cite{eisenberg1973nature}. Similarly, pH variation significantly influences transient dynamics but offers a limited range of tunable states \cite{mcclintic2023heterosynaptic}. Membrane type and temperature can introduce diverse effects on alamethicin response \cite{najem2018memristive}, but implementing a system with multiple materials or temperatures becomes considerably more labor-intensive.

\subsubsection{Characterizing input nonlinearities}
To determine the impact of alamethicin concentration on the memory and nonlinearity of the memristor, we subjected five memristors, each with a different alamethicin concentration ranging from 1 to 3 \textmu M, to various voltage inputs and analyzed their current responses. We used the resulting experimental data to obtain parameters for the model equations describing an alamethicin-based memristor \cite{najem2018memristive} using an elaborate fitting routine, as described in the Methods section and the Supporting Information (SI). First, we examined the steady-state output of each memristor in response to a slow-voltage ramp (2 mHz, 0 to 170 mV) such that the transient dynamics are not detectable. Figure \ref{fig:fig2}a shows the quasi-steady state response for the five alamethicin concentrations. Our observations reveal that higher concentrations of alamethicin lead to membrane insertion at lower voltage thresholds, which agrees with the literature\cite{okazaki2003ion,eisenberg1973nature}. For instance, with the lowest concentration, we saw insertion occur at around 120 mV while at the highest concentration insertion occurs around 80 mV. Furthermore, the steady-state number of inserted pores, $N_{ss}$, varies exponentially with the input voltage as shown in Figure \ref{fig:fig2}a and described in the model (see Section 2.1 in the SI). We used this data to obtain fitted values for both the number of open pores at 0 V, $N_{0}$, and the voltage required to drive an \textit{e}-fold increase in membrane permeability, $V_e$, as described in the SI. Based on our fitted parameters presented in Table S1, we conclude that the exponential rate of increase in steady-state response is the same for all concentrations once insertion begins. This finding aligns with existing literature, which suggests that the rate of exponential increase with voltage is solely determined by constants associated with the dipole moment of alamethicin, rather than the concentration of alamethicin in solution \cite{okazaki2003ion}. However, each concentration of alamethicin results in a distinctive time constant and steady-state pore concentration value, ensuring a unique nonlinear transformation of input voltages with identical pulse width and amplitude. A similar effect can be seen for changing voltage offset while keeping the same concentration (Figure \ref{fig:fig2}b and \ref{fig:fig2}c).

\subsubsection{Characterizing memory timescales}
\begin{figure}
  \includegraphics[width=\linewidth]{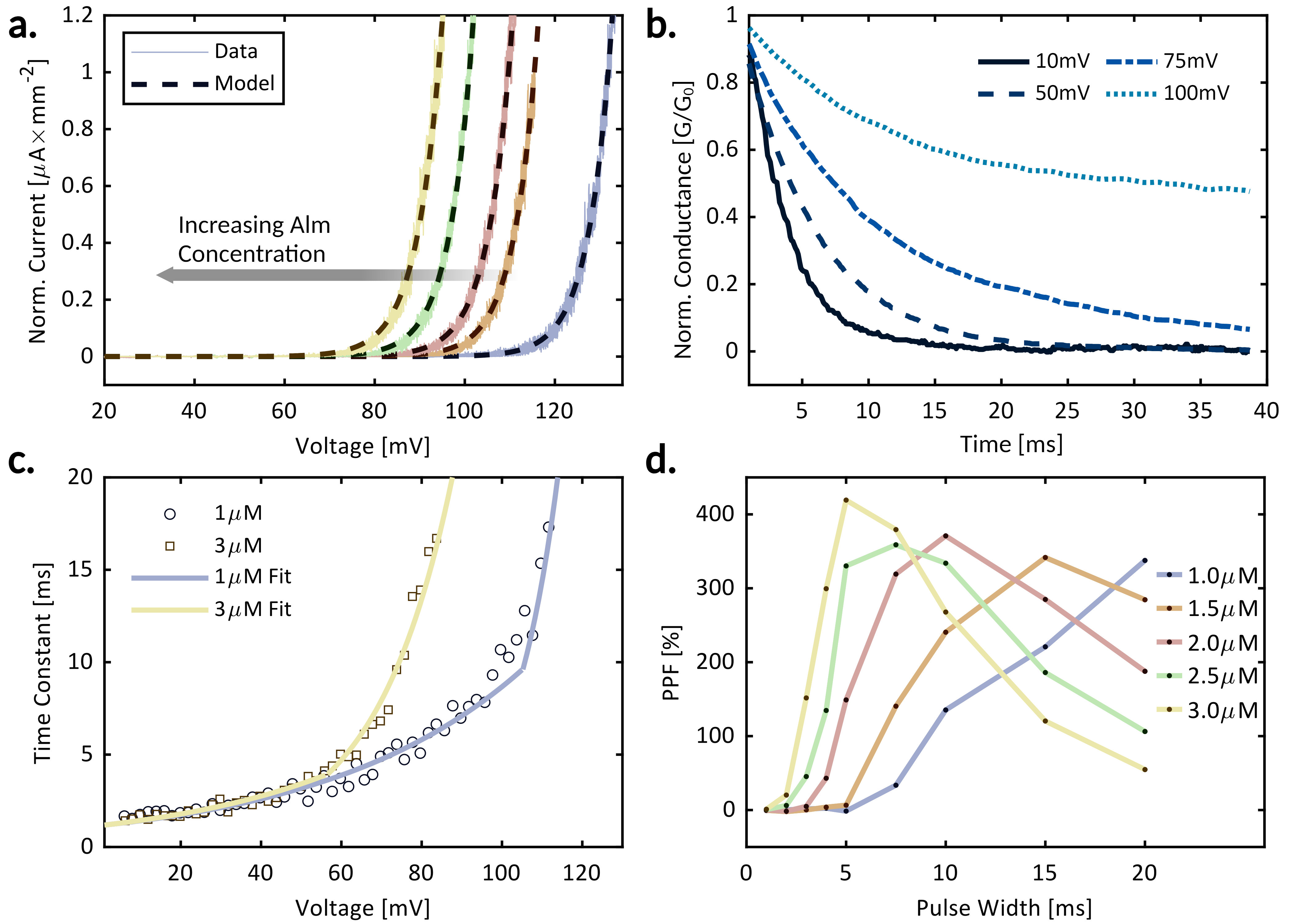}
  \caption{The temporal characteristics and the memristance nonlinear relationship with voltage of ion alamethicin memristors. (a)The quasi-steady state current output of alamethicin memristors with different concentrations to voltage input. We varied the voltage at a rate of 2 mHz such that no hysteresis was observed. Alamethicin concentrations from left to right are 3 $\mu$M, 2.5 $\mu$M, 2 $\mu$M, 1.5 $\mu$M, and 1 $\mu$M. We notice that as concentration increases, the exponential response occurs at lower voltages. (b) A 3-$\mu$M memristor is subjected to a DC input of 160 mV until a steady state is reached. The voltage input is then decreased to one of four voltage levels: 100, 75, 50, and 10 mV. The corresponding normalized conductance for each held offset is plotted, where the time constant of decay can be seen to increase with increased voltage. (c) The effect of voltage and alamethicin concentration on the time constant of the response is shown to be exponential for each concentration, with a dramatic change in exponential rate near the insertion threshold of that concentration. (d) The PPF of unbiased memristors with five different concentrations of alamethicin. Each memristor responds most prominently to a narrow range of stimulus times. All pulses were applied with an IPI of 5 ms and amplitude of 170 mV.}
  \label{fig:fig2}  
\end{figure}

Next, we investigated the impact of bias voltage and concentration on the short-term memory of the devices. The memory behavior of a dynamic memristor often emerges from the disparity between the rates of potentiation and decay. This phenomenon is best represented by the pinched hysteresis observed in the I-V plane, as initially elucidated by Leon Chua\cite{chua1971memristor}. With this concept in mind, we conducted voltage sweeps at a frequency of 200 mHz across all memristors and recorded their corresponding current outputs. Our findings revealed a pinched hysteresis pattern across the I-V plane for all devices, indicating the presence of dynamic memory (Figure S1). Our observations indicate that the hysteresis is more significant for devices with large alamethicin concentrations. To quantify these observations (i.e., the impact of concentration and voltage on memory), we recorded the conductance decay of each device and fit the data to our model to extract parameters. Figure \ref{fig:fig2}b shows the decay of a 3-$\mu$M-memristor as we reduced the applied voltage from 160 mV to one of four voltage levels (100 mV, 75 mV, 50 mV, and 10 mV). It is important to clarify that the voltages indicated on the graph are the voltages that the waveform drops to. For this study, we normalized the resulting device conductance at each new voltage level to the initial conductance (at 160 mV) to ensure that the decay rates were compared appropriately. We found that these responses are strongly exponential, indicating that the first-order differential equation \cite{eisenberg1973nature} describing this system is approximately linear. Therefore, we fit these responses to an exponential decay function in time. We then mapped the time constants of these fits to the voltage at which they occur for each concentration, creating functional dependencies of the time constants to voltage. Figure \ref{fig:fig2}c shows the voltage-dependent change in the time constant for two of the concentrations. We notice that for sub-threshold voltages, the time constants increase with voltage at approximately the same exponential rate, regardless of concentration. However, after the insertion threshold for each concentration is passed, the exponential rate of increase in time constant with respect to voltage becomes larger. Therefore, we conclude that the time constants display a functional dependency on concentration based on the insertion threshold. 
To delve deeper into this phenomenon, we subjected all devices to voltage pulses and recorded the resulting currents. These results enabled us to explore their performance in terms of PPF which describes the short-term increase in synaptic conductivity following a pulse \cite{zucker1989short}. As previously discussed, memory retention in these devices predominantly hinges on the discrepancy between the rates of potentiation and decay. Consequently, when a second pulse follows the first pulse after a specific interval (i.e., the inter-pulse interval (IPI)), the synapse may experience heightened potentiation due to insufficient return to the resting state between pulses. Mathematically, PPF is the ratio of the difference between the maximum potentiation during the first and second pulses to the maximum potentiation during the first pulse. PPF can be influenced by both the IPI and the pulse width (PW). In Figure \ref{fig:fig1}d, we illustrate the PPF profiles for the five devices against the IPI, with a consistent pulse width and applied voltage (5 ms, 170 mV). These results display a trend of higher PPF values for greater alamethicin concentration, at any given pulse width. 
The inset diagram portrays two instances of voltage pulses and their corresponding blue current responses, showing approximately 100\% growth. Our findings also reveal an exponential decline in the degree of potentiation as the IPI increases for all devices. Importantly, the behavior of PPF for a specific memristor can be influenced by the pulse width if the initial potentiation response is non-linear. The visual representation in Figure S3 illustrates the diverse variations of PPF in relation to both PW and IPI. The behavior observed with small pulse widths is of particular interest, where higher alamethicin concentrations exhibit the most pronounced potentiation enhancement. However, as the pulse width increases, lower alamethicin concentrations exhibit progressively higher PPF. Ultimately, a significant inversion occurs, with the lowest concentration manifesting the highest PPF during extended pulse widths. This phenomenon is illustrated in Figure \ref{fig:fig2}d, where a 5-ms PW results in an increase in PPF with concentration, while a 20-ms PW leads to a decrease in PPF with concentration. It is crucial to emphasize that all administered pulses maintained a constant voltage of 170 mV in a square waveform. These findings robustly confirm that our alamethicin-based devices exhibit distinct memory timescales influenced by their concentration and the applied voltage. Additionally, aside from the sensitivity of PPF to pulse width, an increase in IPI results in a decrease in PPF for all concentrations, as depicted in Figure \ref{fig:fig2}d. Note that the PPF values presented in both Figure \ref{fig:fig1}d and Figure \ref{fig:fig2}d are calculated based on positive voltage pulses with an off-voltage of 0 V. Changing the off-voltage to a higher positive value would affect the relative PPF due to the change in the decay rate induced by the voltage, as illustrated in Figure \ref{fig:fig2}b. Thus, we conclude that each memristor exhibits a more pronounced response to specific temporal input ranges. Furthermore, when exposed to an identical voltage waveform simultaneously, The response of each memrisotr is triggered by different temporal features within the waveform, resulting in a diverse array of responses. Additionally, our device's unique exponential voltage dependence of the time constant enables the extraction of multiple timescale features from synchronous temporal inputs. By varying the offset voltage applied to multiple identical devices in a reservoir, each memristor is able respond to features of different timescales most prominently.

\subsection{Multi-device memristive reservoirs for temporal processing and classification}

To demonstrate the enhanced performance of the diverse memristor reservoir with multiple memory timescales and nonlinear dynamics, we consider two tasks: a dynamic model prediction task and a time series classification task. In the previous section, we showed that alamethicin memristors can exhibit different memory timescales by varying the alamethicin concentrations or biasing a device with a DC voltage. For the dynamic model prediction task, we constructed our reservoir using five distinct memristors, each unbiased and with a different alamethicin concentration, resulting in unique memory timescales. We demonstrate that each memristor produces a linearly independent response to the input data. However, a linear combination of all responses outperforms any single device by extracting more temporal features. This task is well-suited for changes in alamethicin concentration, as the input waveform can be encoded to a common voltage suitable for all memristors. For the time series classification task, three memristors of 3 $\mu$M alamethicin concentration were each biased with a different DC voltage offset. With this approach, we sent the same time series to each memristor and extracted virtual nodes at regular intervals, which resulted in different responses due to the voltage biases. While it is possible to solve this task by varying alamethicin concentrations, we opted for the voltage offset implementation to demonstrate its similar effectiveness. Additionally, lower alamethicin concentration devices exhibited higher noise-to-signal ratios that hindered the results. Regardless of the approach to vary the memory timescale, the objective of our study is to validate our original hypothesis that multi-device memristive reservoirs with various memory timescales and nonlinearity deliver more accurate predictions and classifications while synchronously encoding the input signal only once.

\subsubsection{Predicting a second-order nonlinear dynamical system}
\begin{figure}
  \includegraphics[width=\linewidth]{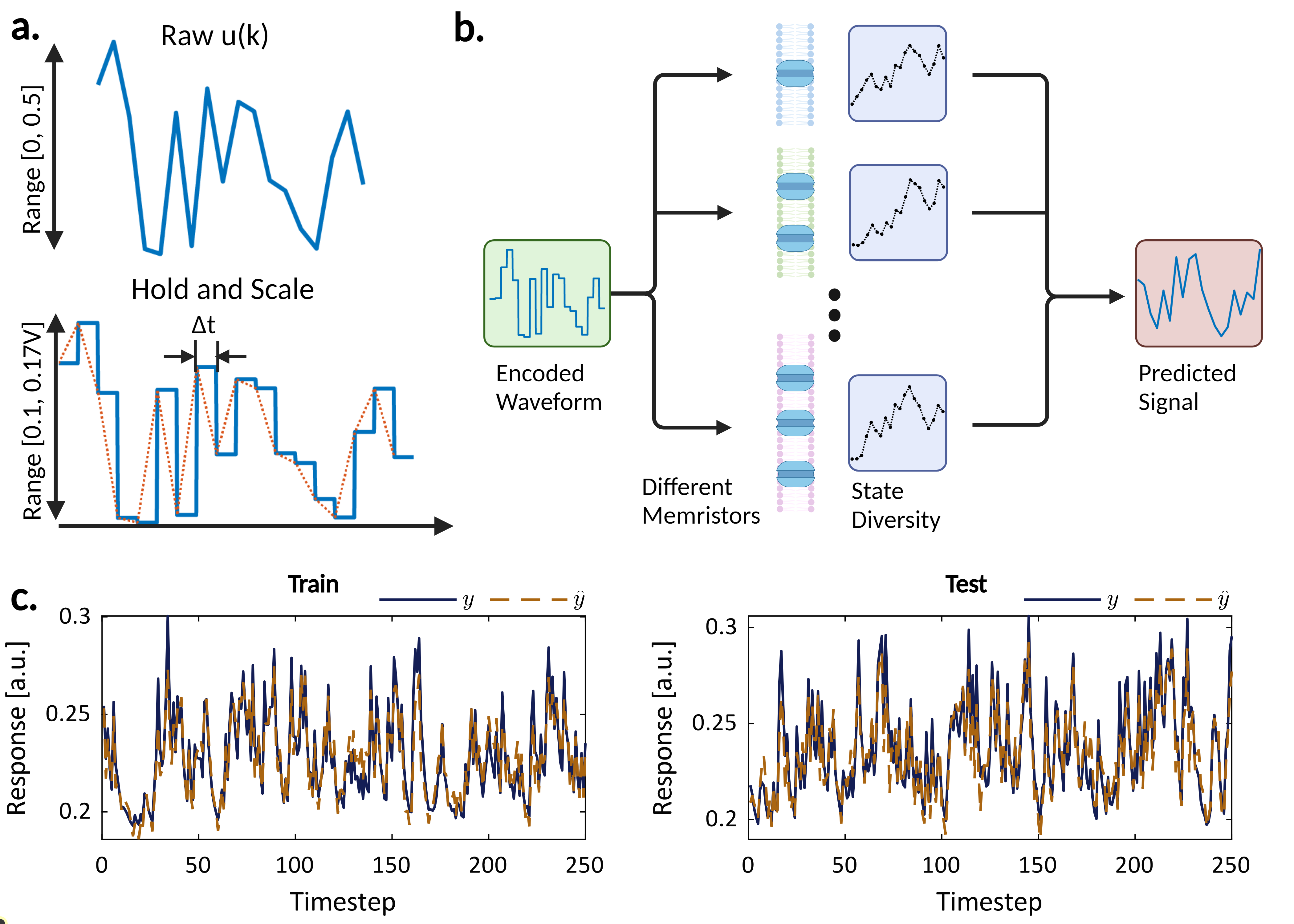}
  \caption{The process and results of the second order task using a diverse memristor-based reservoir. (a) The raw signal, $u(k)$, ranges from 0 to 0.5 for discrete time steps. We transformed it into $V(t)$, a voltage signal that is linearly correlated with $u(k)$ and held for a time $\Delta t$, which was 3 ms for all cases we tested in this study. (b) A diagram showing the signal voltage waveform that was sent to all memristors. We recorded the conductance of each memristor at each held interval and used the values as states for the reservoir. These states were then multiplied by the trained weights to reconstruct the predicted signal. (c) The training and testing waveforms for the second-order task. Displayed is the best result for a reservoir of five memristors with NMSE of $1.2\times 10^{-3}$ and $1.5\times 10^{-3}$ for training and testing, respectively. In these plots, $y$ is the true value, and $\hat{y}$ is the predicted value.}
  \label{fig:fig3}
\end{figure}

The second-order nonlinear dynamical system (SONDS) task, is a benchmark prediction problem commonly employed in the field of physical reservoir computing \cite{du2017reservoir,maraj2023sensory,taniguchi2022spintronic}. This task requires both nonlinearity and short-term memory to be successfully predicted using a memristor-based reservoir \cite{du2017reservoir,maraj2023sensory,taniguchi2022spintronic}. Previous attempts to solve this task with memristor reservoirs have resorted to utilizing different input encodings to manipulate the timescales of the input data, resulting in diverse responses from identical memristors \cite{du2017reservoir, maraj2023sensory}. In contrast, for this task, we maintain a fixed encoding while employing memristors with varying alamethicin concentrations. This approach enhances dimensionality by introducing diverse nonlinear and short-term memory dynamics. The primary objective of this task is to train a reservoir computing model with empirical data to predict output, $y(k)$, based only on input, $u(k)$, with no prior knowledge of the underlying transfer equation defined by:
\begin{equation}
\label{6}
y(k)=0.4y(k-1)+0.4y(k-1)y(k-2)+0.6u^3(k)+0.1
\end{equation}
To this end, we randomly generated two sequences of values for $u_1(k)$ and $u_2(k)$, uniformly distributed from $[0, 0.5]$ as inputs to the reservoir to train and test the reservoir computer, respectively. Next, we trained the memristor reservoir to produce a prediction, $\hat{y}_1(k)$, of $y_1(k)$ by encoding the sequence $u_1(k)$ to a voltage waveform that we sent to all memristors. The sequences $u_1(k)$ and $u_2(k)$ were linearly scaled and encoded into voltage waveforms, $V_1(t)$ and $V_2(t)$ based on the following equation:
\begin{equation}
    V(k\Delta t:k\Delta t+\Delta t)=\gamma u(k)+\delta
\end{equation}
where $\Delta t$ is the held time, $\gamma$ is the linear scale parameter, and $\delta$ is the offset parameter. We later sent the generated waveform to all memristors and recorded their output current, as described in the Methods section. We sampled the conductance of each memristor once per encoded point to obtain the full state of the reservoir at each time step $k$. Figure \ref{fig:fig3} describes the voltage encoding process of \textbf{Equation 2}, in which the value $u(k)$ was multiplied and offset by constant values $\gamma$ and $\delta$ respectively. The resulting value was held for a time, $\Delta t$. The parameters $\Delta t$, $\gamma$, and $\delta$ were determined by a simulated grid search, conducted using the model obtained from experiments in the previous section (See SI discussion 4).

The final step of the training phase involved utilizing linear regression to determine the best-fit weights vector, which mapped the reservoir states (i.e., conductances) to $y_1(k)$ using 250 values of $u_1(k)$. A new sequence, $u_2(k)$ was encoded and fed into the reservoir to assess the quality of the training phase. Predicting $\hat{y}_2(k)$ was accomplished by computing the vector-matrix product of the weights vector and the new reservoir state matrix. The comparison between the predicted values, $\hat{y}_2(k)$ and $y_2(k)$, (values derived from Equation 1) constitutes the "testing phase" which also involved 250 values of $u_2(k)$.

In this study, we found that only five memristors were needed to comprise the reservoir layer, each with a different concentration of alamethicin (the same concentrations described in the previous section). The required number of memristors was determined by evaluating the prediction accuracy of the reservoir in both the training and testing phases. Evaluation of the reservoir was achieved by comparing normalized mean square errors (NMSEs) between reservoirs (as described in the Methods section). Of course, if the NMSE values are not small enough, additional memristors of increasing alamethicin concentration can be added to the reservoir, and further diversify the state space. 

The results we obtained using the five memristors and the optimized input encoding resulted in a training NMSE value of $1.3\times 10^{-3}$ and testing NMSE value of $1.5\times 10^{-3}$. These values are lower than experimental results found for 90 metal-oxide memristors run with various input encoding methods employed simultaneously which had a testing NMSE of $3.13\times 10^{-3}$ \cite{du2017reservoir}. We also show that noiseless simulations of the five memristors offer even lower testing NMSE of $2.18\times 10^{-4}$ (Figure S5), which is lower than simulations of 90 ion-channel-based memristors which achieved less than $9.4\times 10^{-4}$ \cite{maraj2023sensory}. In addition to the improvement in accuracy, the reduction in size of the reservoir layer from 90 devices to 5, reduces the number of trained weights in the readout layer from 91 to 6.

These results not only validate our main hypothesis, but also demonstrates the efficacy of our approach, which leads to better prediction results with significantly fewer memristors and less pre-processing required, irrespective of the memristor type. These results suggest that proper implementation of this approach with any type of volatile memristor capable of tunable dynamics will lead to less energy consumption and a smaller footprint.  

\subsubsection{Neural Activity Classification} 
To further showcase the versatility of tunable memristors and support the main hypothesis of this study, we consider a task that requires synchronization of data streams. This reservoir computing classification task was created by Zhu \textit{et al.} \cite{zhu2020memristor} and entails distinguishing various emulated patterns of neural activity into four common classes: tonic, bursting, adapting, and irregular \cite{zhu2020memristor,john2022reconfigurable}. These emulated patterns of neural activity were sent as voltage waveforms to a single dynamic memristor whose conductance was sampled at regular intervals over the time span of the sequence. The conductance of the memristor changed with increased or decreased stimulation from the neural activity, encoding the  data into virtual reservoir nodes at evenly spaced, sampled points \cite{zhu2020memristor}. Zhu \textit{et al.} noted that neural signals are often composed of multiple timescale features, and therefore, memristor reservoirs with various timescale devices may improve the classification ability for this task as well as more complex tasks \cite{zhu2020memristor}.

\begin{figure}
\includegraphics[width=\linewidth]{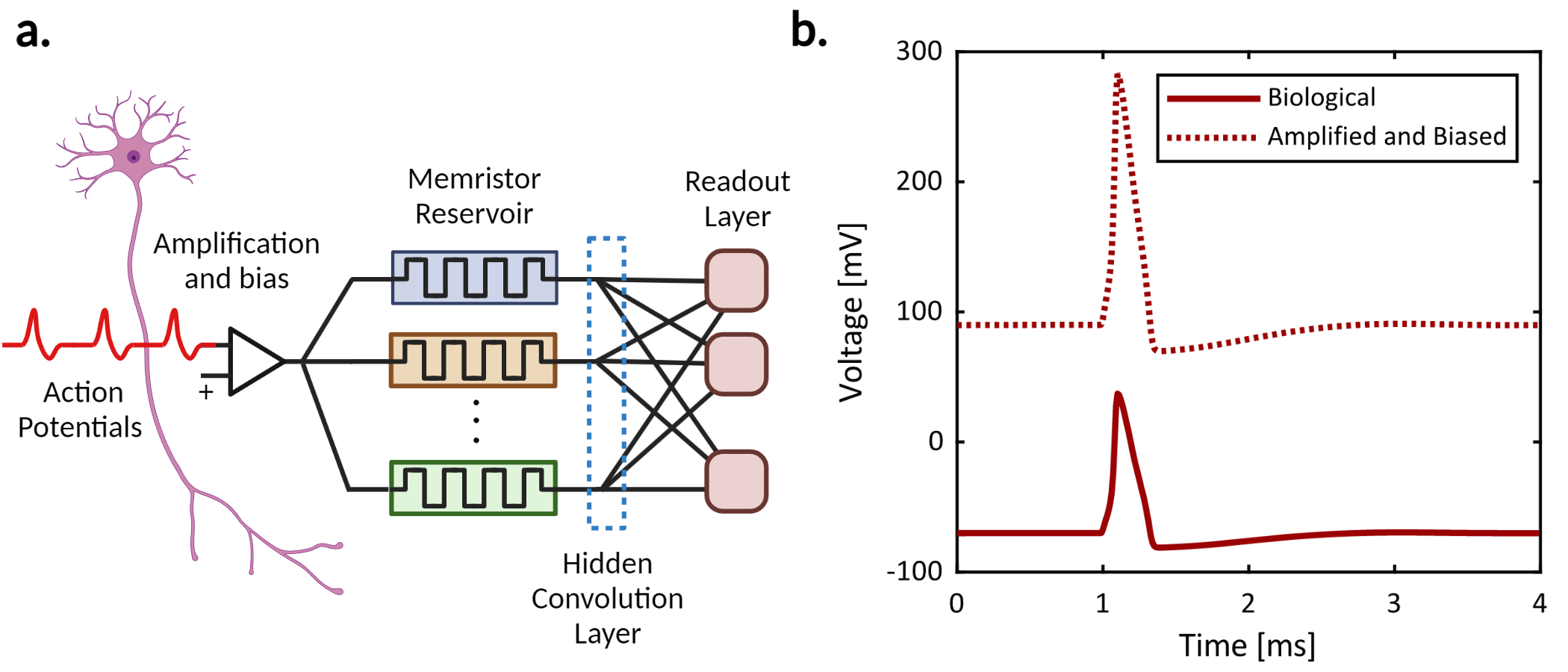}
\caption{(a) A diagram illustrating the neural activity classification task. Biological signals are biased and amplified before being sent to one or more memristors. The sampled states of the memristors form the reservoir layer. Optionally, a hidden 2-D convolutional layer separates the reservoir from the readout layer, connecting adjacent nodes and reducing the number of trained weights. (b) A comparison of an approximate biological action potential to the scaled and biased action potential that is transmitted to the memristors. Importantly, the modified waveform does not alter the timescale of the pattern, allowing real-time signal processing.}
\label{fig:fig4}
\end{figure}

The published results, which achieved an accuracy of 88-92.5 \% \cite{zhu2020memristor}, could be further enhanced by utilizing the same waveforms with memristors featuring diverse dynamics, thus creating a larger reservoir layer. To achieve this improvement, we implemented a multi-memristor reservoir designed to introduce variability into the reservoir layer (as depicted in Figure \ref{fig:fig4}a). Specifically, we manipulated the time constant of multiple memristors through changes in DC voltage offset. Subsequently, we exposed each of these devices to the same input patterns. We analyzed the reservoir outputs in two distinct ways and compared the results. First, the outputs were directed to a fully connected (FC) readout layer. As a point of comparison, we also routed the output of the reservoir into a bilayer convolutional neural network (CNN) equipped with a hidden 2-D convolution layer. This convolution layer enabled the coupling of virtual nodes originating from multiple memristors, each containing different timescale information, which would otherwise remain independent of each other. The accuracy substantially improved with the addition of a 2-D convolution layer (Figure \ref{fig:fig6}), and the number of trained weights decreased (Figure S8).


\begin{figure}
  \includegraphics[width=\linewidth]{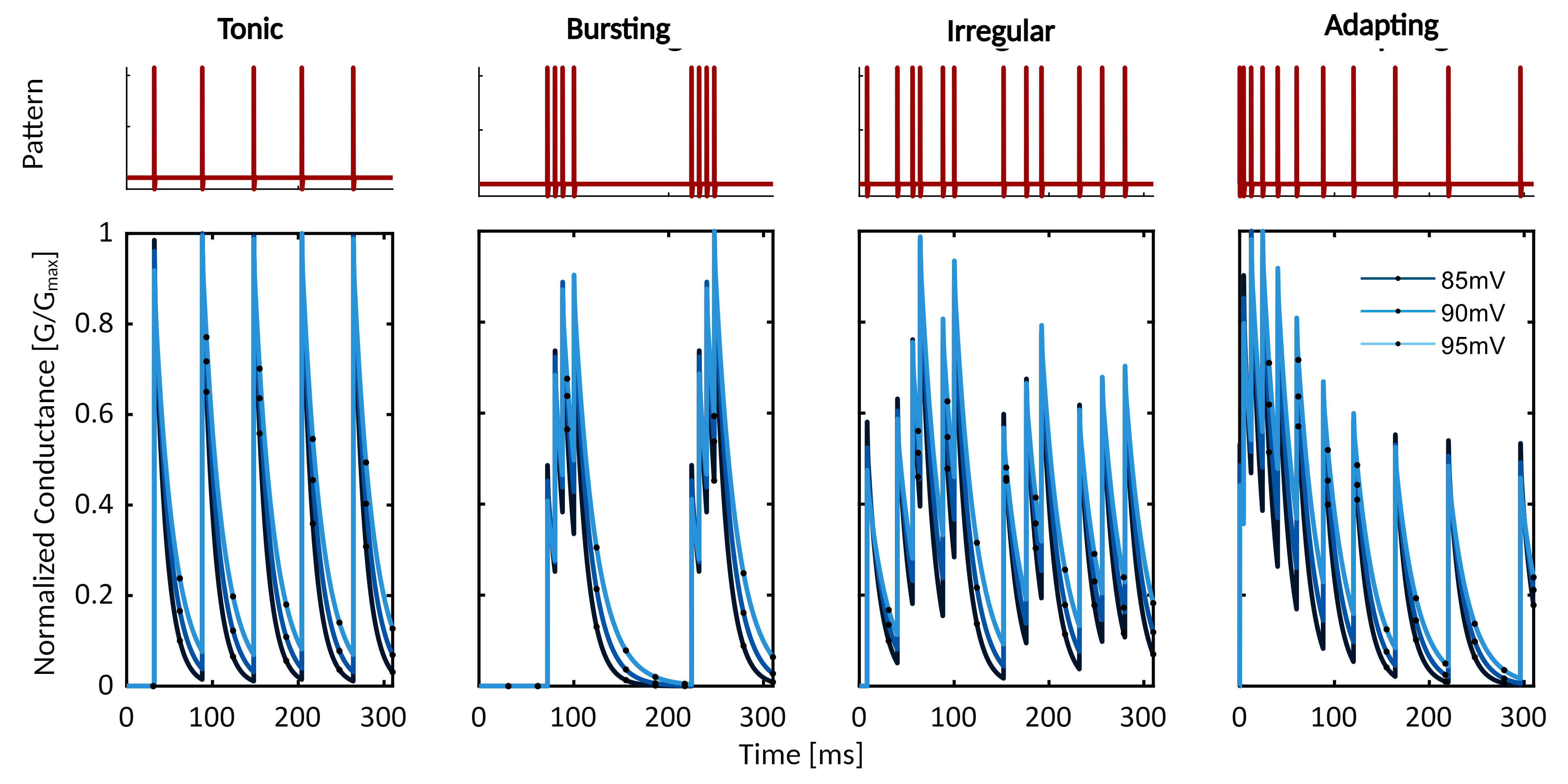}
  \caption{Simulations illustrating the four classifications of neural activity data. Each time sample is 620 ms in length, but here, we display half a sample for each to provide detail. In all cases, the same data pattern was sent to the three memristors with different DC voltage offsets, and we observed slight variations in their normalized responses. On each plot, the marked points represent the sampled points for each memristor, serving as virtual nodes of the reservoir. The unique variations from one point to another are distinctive for each memristor, indicating their ability to extract different temporal features from the input data stream.}
  \label{fig:fig5}
\end{figure}

\begin{figure}
  \includegraphics[width=\linewidth]{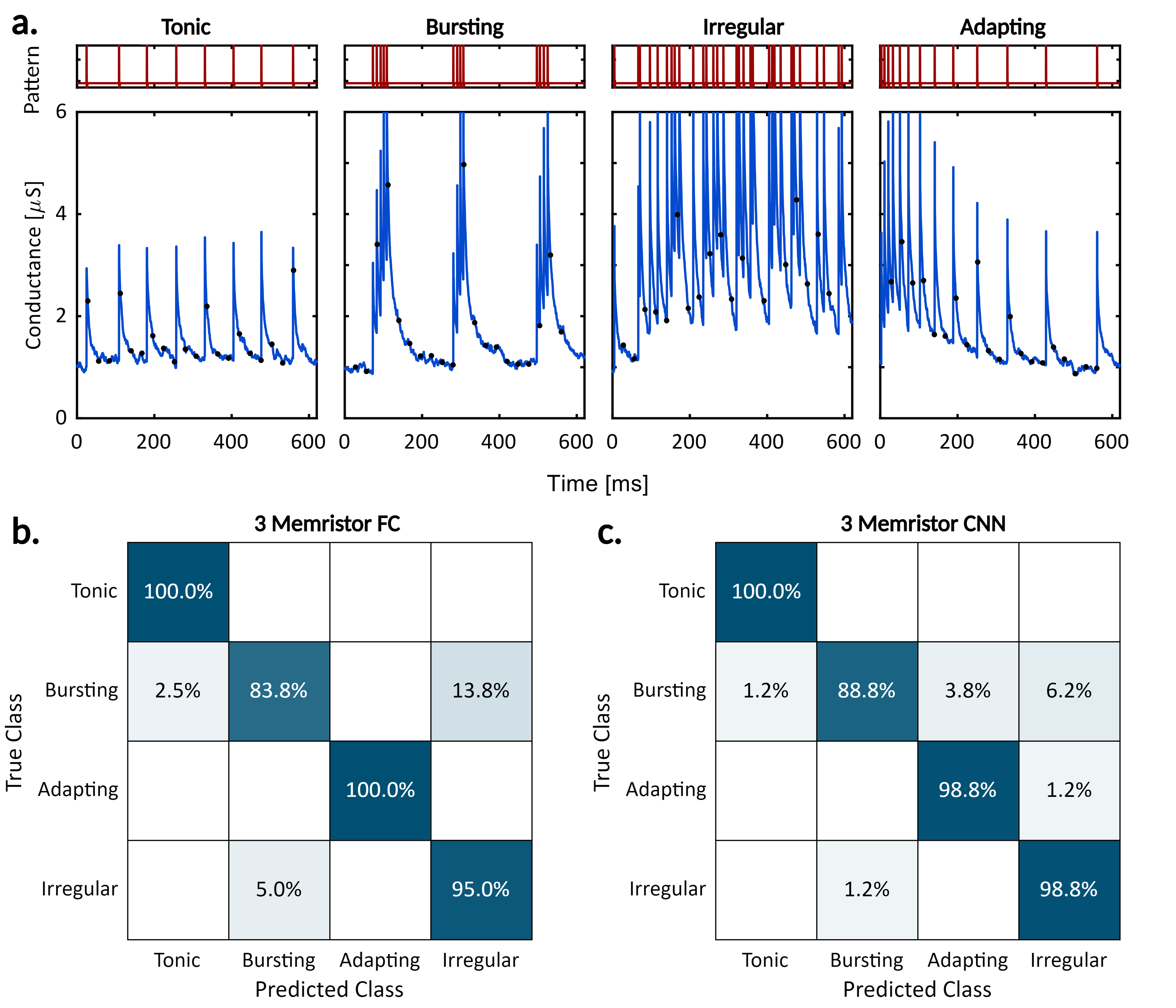}
  \caption{(a) Example data for the four classes of neural activity. The top row displays the voltage patterns of neural activity, while the bottom row presents the complete response of a $3\mu$M memristor to the respective waveforms sampled at 10 kHz. The marked points in the bottom row represent the responses sampled at 32.2 Hz, equivalent to 20 samples per activity pattern. (b-c) Confusion matrices show the experimental results for a 3-memristor FC and 3-memristor CNN with various DC offsets, respectively. In the FC case, the overall accuracy reached 93.4\%, whereas the CNN case achieved an overall accuracy of 96.5\%.}
  \label{fig:fig6}
\end{figure}

First, we conducted a simulation using three memristors to assess the feasibility of this task. We implemented the memristor reservoir by employing three parallel $3 \mu M$ alamethicin memristors. Similar to the SONDS task, we encoded the data by scaling and offsetting the initial signal. To introduce independent dynamics for the three memristors, each device recieved a distinct signal voltage offset of 85 mV, 90 mV, and 95 mV, while they all shared the same signal scaling of $1.8\times$ (as depicted in Figure \ref{fig:fig4}b). The alterations in rise and decay rates, governed by the exponential relationship between the DC offset and the time constant, resulted in changes in the sampled virtual nodes. For instance, a slower decay rate could lead to facilitation even when pulses were temporally distant, whereas a faster decay rate might only show facilitation when multiple pulses occurred in quick succession. Consequently, each memristor responded the strongest to different temporal features in the incoming data. As a result, no single memristor captured the complete response of the incoming waveform, but by combining multiple memristors into a single reservoir, more temporal features could be included in the classification task. Figure \ref{fig:fig5} displays the complete simulated response of the three memristors to examples of each classification, along with the virtual nodes. These nodes exhibited similar yet independent variations from one another. The independent variance of virtual nodes is crucial for reservoir performance because linearly dependent nodes do not increase the dimensionality of the reservoir state matrix. We assessed and compared the results of reservoir layers formed from both singular devices, as well ass all three devices used concurrently. Furthermore, by adjusting the sampling rate, we obtained various numbers of virtual nodes for each sample. In general, increasing the number of virtual nodes improved the accuracy of the classification task, although it required training more parameters. Finally, reservoirs with three memristors and CNN output layers outperformed reservoirs with a single memristor and FC output layers, all while using fewer trained parameters (as shown in Figure S8). These noiseless simulation results yielded a testing accuracy of 94.68\% in the best three-memristor FC case and 99.06\% in the three-memristors CNN case (as illustrated in Figure S9). 

Experimental results were obtained for the same parameters as the simulated results for memristors, scalings, and offsets. We sampled the memristors at various rates to obtain accuracy values for different numbers of virtual nodes. For both the 1-memristor and 3-memristor reservoirs, we found that more than 20 virtual nodes per sample gave quickly diminishing returns on accuracy in the FC output layer implementation. We observed $93.4\%$ accuracy using 20 virtual nodes per memristor, and only a slight increase to $94.0\%$ using 155 virtual nodes per memristor. Additionally, the number of trained weights from 20 to 155 virtual nodes increased from 244 to 1864. Nonetheless, we increased the accuracy and reduced the number of trained weights simultaneously by implementing a hidden 2-D convolution layer. The convolution layer received the output of the reservoir layer, which was shaped $(3,20)$ corresponding to the three memristors and the 20 virtual nodes per memristor. We optimized the kernel shape of the 2-D convolution to $(3,9)$, which resulted in an improved $96.5\%$ experimental testing accuracy with only 80 trained parameters (Figure \ref{fig:fig6}). Compared with the previous work on this task with a single memristor, the error rate was reduced from $7.5\%$ to $3.5\%$ (less than half), while simultaneously reducing the number of trained parameters from 128 to 80 \cite{zhu2020memristor}. 

\section{Discussion and Conclusion}
We demonstrated that the short-term memory and nonlinear dynamics of alamethicin ion-channel-based memristors can be tuned by varying alamethicin concentration or applying a voltage bias. These findings highlight the effectiveness of varying the dynamic properties of memristors within a reservoir layer to enhance the richness of the state space, resulting in improved predictive and classification capabilities using the same encoding of the input data. Compared to alternative methods for expanding the state space, such as varying signal timings or introducing random variations in device parameters, our approach offers the advantage of enabling real-time deployment of dynamic reservoirs. The presented methodology ensures that the start and end times of data input are synchronized for all memristors, allowing for controlled variation in their responses. Moreover, our method requires minimal pre-processing, as the input data in both tasks were simply scaled and offset to voltage waveforms. Additionally, we accomplished these results using ion-channel-based memristors, which operate with low voltage and current requirements, making them suitable for direct connection to biological signals. We note that for testing purposes, we fabricated these devices as droplet interface bilayers (Methods Section). In principle, the functional unit of these memristors consists of ion channels in lipid bilayers at the nanometer scale. We anticipate that the macroscopic functionality demonstrated here can be scaled down to much smaller devices, which will be a focal point of future work. Lastly, it is worth mentioning that the architecture of parallel memristor reservoirs consists of self-recurrent linear nodes, with no connections between the nodes in these reservoirs. In the future, we will concentrate on integrating these biomolecular memristive devices into more complex networks, where connections between nodes will drive richer dynamics. Introducing dynamical nonlinearities into the memristor update process, rather than solely input nonlinearities, will enable these devices to achieve higher accuracy in tasks that demand more nonlinearity than memory.


\section{Methods Section}
\subsubsection{Device Characterization}
The various waveforms we used to characterize the properties of memristors were selected to fit for various parameters in an existing model of alamethicin insertion \cite{najem2018memristive}:
\begin{equation}
    \label{5}
    \frac{dN_{a}}{dt}=\frac{1}{\tau_{0}e^\frac{V}{V_{\tau}}}(N_{0}e^\frac{V}{V_{e}}-N_{a})
\end{equation}
where $N_a$, $N_0$, $V_e$, $\tau_0$, and $V_\tau$ are the number of pores per unit area, the number of pores open at zero applied voltage, the voltage required for an \textit{e}-fold increase in the steady-steady state number of pores, the time constant at zero applied voltage, and the voltage required for an \textit{e}-fold increase in time constant\cite{najem2018memristive,okazaki2003ion,eisenberg1973nature}. The multiple voltage waveforms we used to characterize each memristor were designed to determine a particular set of parameters while remaining independent of those not under test. The first of these waveforms is a slow voltage sweep at a rate of 2 mV/s, which ensures that any transient effects of the devices are ignored. With each voltage increment, the devices reach their steady-state values. This allows for the fitting of model parameters $N_0$ and $V_e$ for each memristor as these parameters are defined based on the steady-state values of conductance for each memristor as described by the following equation:
\begin{equation}
    N_{SS}(V)=N_0e^{V/V_e}
\end{equation}
where $N_{SS}(V)$ is the steady-state value of pores per area as a function of voltage, as determined by the slow voltage sweep. Furthermore, we applied a second waveform to the memristors, consisting of stepwise voltage drops. Specifically, the voltage was maintained at a constant level of $V_{high}$ for 500 ms, then reduced to a variable lower voltage, $V_{low}$, which is held steady for 300 ms before returning to $V_{high}$, followed by the next $V_{low}$ (see Figure S4). The value of $V_{high}$ was determined for each concentration to be the voltage at which the memristor conducted 190 nA of current and ranged from 95 mV at 3 $\mu$M to 133.6 mV at 1 $\mu$M. The slow decay of conductance in response to this step voltage change was then fit to an exponential decay such that:
\begin{equation}
    G(t)=G_0e^{-t/\tau}
\end{equation}
where $G(t)$ is the change in conductance over time $G_0$ is the initial conductance, and $\tau$ is the time constant of decay that is being fit for(Figure \ref{fig:fig2}c). The time constant, $\tau$, was subsequently fitted as an exponential function of voltage, with each $\tau(V)$ being fitted to the corresponding $V_{low}$.
\begin{equation}
    \tau(V)=\tau_{0}e^{V/V_{\tau}}
\end{equation}
This fit was then split into two sections as our observations concluded that the $\tau$ of each memristor exhibits different dynamics at voltages below the insertion threshold of alamethicin. $V_{threshold}$ was determined by the insertion threshold of alamethicin, which is the voltage for single alamethicin channels to begin to form in the membrane. Further details on the model (\textbf{Equation 3, 4, 5} and \textbf{6}) and fitting routines can be found in the SI.

\subsection{Material Preparation}
A stock aqueous solution of salt was first prepared of potassium chloride (1 M) and MOPS buffer (10 mM) in deionized (DI) water, with a measured pH of 5.8. Separately, a chloroform solution containing 1,2-diphytanoyl-sn-glycero-3-phosphocholine (DPhPC) lipids at 25 mg/mL concentration (purchased from Avanti Polar Lipids, Inc) was evaporated and the remaining lipids were left under vacuum for 1 hour to ensure all chloroform was removed. Next, the stock aqueous solution was added such that the concentration of DPhPC was 2 mg/mL and the resulting lipid solution went through five freeze-thaw cycles. Then, the solution was extruded through a 0.1 $\mu$m pore track-etched membrane (Whatman\textsuperscript{TM}) for eleven passes. Finally, the lipid solution was allocated into five vials. Before every experiment, the alamethicin peptides in ethanol solution (2.5 mg/mL F50 from AG Scientific, Inc.) were mixed in each of the lipid solutions to a final concentration of 1 $\mu$M, 1.5 $\mu$M, 2 $\mu$M, 2.5 $\mu$M, and 3 $mu$M, respectively.

\subsection{Experimental Setup}
All electrical experiments were conducted using droplet interface bilayers (DIBs) in the "hanging-drop" configurations, as described in detail by Najem \textit{et al.}\cite{najem2019assembly} (Figure S12). Silver/silver chloride (Ag/AgCl) electrodes were dipped into molten agarose to thinly coat them with the hydrophilic gel. The wires were then suspended into a reservoir of hexadecane oil (Sigma-Aldrich\textsuperscript{\textregistered}), and positioned on an upright microscope (IX73, Olympus) for observation. Droplets (200 nL) of each alamethicin concentration were pipetted onto the wires and brought together to form bilayers. Electrical interrogation of the devices was performed using either a patch-clamp amplifier (Axopatch 200b, Molecular Devices) for currents under 200 nA, or a custom-built transimpedance amplifier circuit (Figure S11) for currents above 200 nA. Voltage signals were supplied using a 16-bit data acquisition (NI-9264, National Instruments) at a sampling rate of 10 kHz.

\subsection{Second-Order Non-linear Dynamical System}
For this task, we created voltage waveforms by linearly scaling the input data $u(k)$ from its original range [0, 0.5] into an appropriate range. The scaling parameter $\gamma$, offset parameter $\delta$, and pulse width, were determined by a grid search of simulated memristors (1-3 $\mu$M) using the models with appropriate parameters that we experimentally obtained. We selected 20 values of each parameter and tested all combinations. We simulated the SONDS task in full for each parameter combination and calculated the resulting NMSE to evaluate the quality of the prediction. Low NMSE was found for a range of parameters (Figure S6), and values were picked for $\gamma$ and $\delta$ such that the voltage range was [90, 170] mV and the pulse width was 3 ms.

In total, 300 training and 300 testing data were generated. The testing and training data was passed to each of the memristors and their conductances were measured at the end of each pulse. The first 50 values from each dataset were discarded as they fell within the transient region of the response of the memristor. This data exclusion is a standard practice that has been employed by researchers in other studies, especially for the SONDS task \cite{du2017reservoir,jaeger2004harnessing}. Next, the remaining 250x5 training state matrix was used to train a weights vector using the built-in linear regression function in Matlab. The remaining 250x5 testing state matrix was then multiplied by the weights vector to construct the predicted signal. Finally, the predicted values were compared to the true values obtained from Equation 6, and the normalized mean square (NMSE) was calculated from:
\begin{equation}
    \label{7}
     NMSE=\frac{\sum_{k=1}^N{(y_p(k)-y(k))}^2}{\sum_{k=1}^N{y^2(k)}}
\end{equation}
where $N$ is the total number of tested points, $y_p(k)$ is the predicted value, and $y(k)$ is the true equation value. It is worth noting that most papers that solve the SONDS task use the above definition \cite{du2017reservoir, taniguchi2022spintronic, maraj2023sensory}, however, the more generalized definition of NMSE is given as:
\begin{equation}
    \label{8}
     NMSE=\frac{\sum_{k=1}^N{(y_p(k)-y(k))}^2}{\sum_{k=1}^N{(y(k)-\bar{y}(k))^2}}
\end{equation}
This definition normalizes around the mean value of the signal. It can be seen that \textbf{Equation 7} is simply a special case of \textbf{Equation 8} where the average signal value is zero. Despite this discrepancy, in order to compare against other works, the values reported here use the \textbf{Equation 7} definition.

\subsection{Neural Activity Analysis}
We created voltage waveforms for the neural activity analysis task by generating 400 0.62-s-long patterns for each activity classification. Next, to add randomness, we randomly varied each pulse to within 4 ms of its initial starting point, following the same procedure described by Zhu \textit{et al}. \cite{zhu2020memristor}. In contrast to their approach, we encoded action potential signals into these patterns whenever a pulse was intended to occur, instead of using a square wave approximation of an action potential. Experimentally, the conductance rise and decay of the memristors in response to stimulation by the neural activity must be clearly distinguishable from the system noise, the stochastic effects of the memristor, and thermal and measurement noise. To this end, we amplified the input waveforms and offset them by a DC bias as the native action potentials range from approximately -70 mV to +40 mV \cite{bean2007action}, which would produce low signal-to-noise ratios. This operation changed the range of neural activity data to 70-250 mV, which is well within the operation range of the device and does not require any re-sampling or further pre-processing of the signal. We chose to implement only these simple pre-processing steps because they can be achieved solely through analog circuitry (see Figure S10), making them suitable for real-time experimentation. Subsequently, we transmitted these signals to the memristors and sampled their current output every 0.1 ms, resulting in a total of 6200 sampled states for each pattern. Afterward, we downsampled to a smaller set of variable virtual nodes. We utilized 1280 waveform patterns for training the readout layer, while 320 waveforms were reserved for testing the classification ability of the trained system.  

In the results section, we tested two readout implementations. The first implementation we explored was the FC output layer. For this method, we combined the virtual nodes $n$ from each memristor $m$ into an array of dimensions $(n\times m,1)$. Subsequently, we passed this array through an FC logistic regression layer with a sigmoid activation function (as depicted in Figure S7a, which illustrates a 1-memristor FC configuration). The regression layer was trained using gradient descent within the PyTorch framework, utilizing a binary cross-entropy (BCE) loss function with a learning rate of 0.1. In our second implementation, we combined the $n$ virtual nodes obtained from $m$ different memristors into an array of dimensions $(n, m)$. We then passed this array through a 2-D convolution layer with a kernel of shape $(m, f)$, which generated an output of dimensions $(n - f + 1, 1)$ as shown in Figure S7b. Subsequently, we forwarded this array through a fully connected (FC) layer with a sigmoid activation function. To train this 2-layer readout, we employed gradient descent based on backpropagation within the PyTorch framework, utilizing a binary cross-entropy (BCE) loss function with a learning rate of 0.1.

\medskip
\subsection{Supporting Information:}
Additional information and figures on fitting routines, simulation results, model parameters, and experimental setup (PDF).

\medskip
\subsection{Acknowledgments}
All figures were prepared using biorender.com

\medskip

\bibliography{Library}

\end{document}


\section{Supporting Figures}
\begin{figure} [H]
    \includegraphics{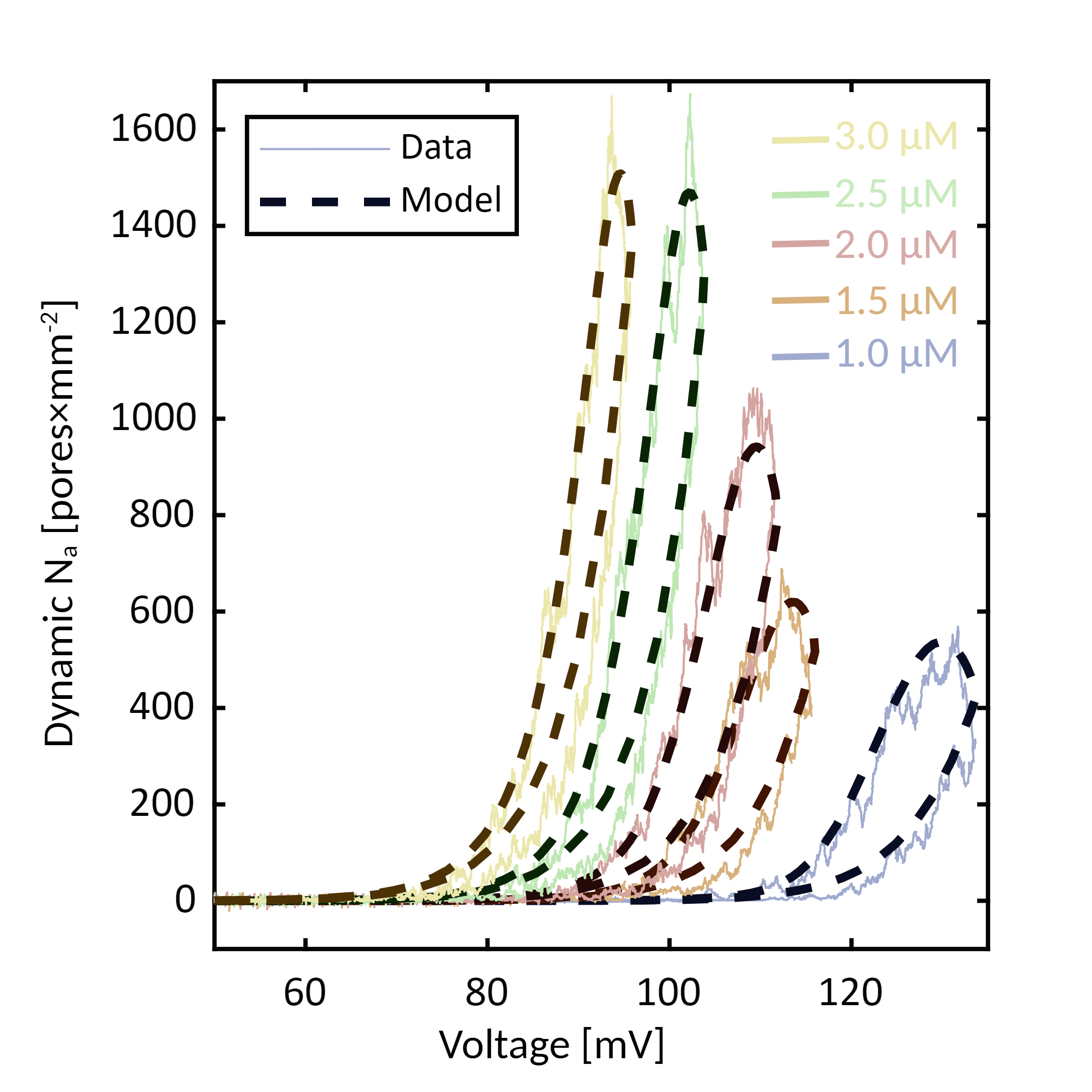}
    \caption{The dynamic response of 5 different concentrations of alamethicin memristors to 200 mHz voltage sweeps. These hysteresis loops confirm that all tested concentrations behave as memristors with varying rates of potentiation and decay. Furthermore, the model's response closely matches the experimental data for all concentrations.}
    
\end{figure}
\begin{figure} [H]
\includegraphics[width=\linewidth]{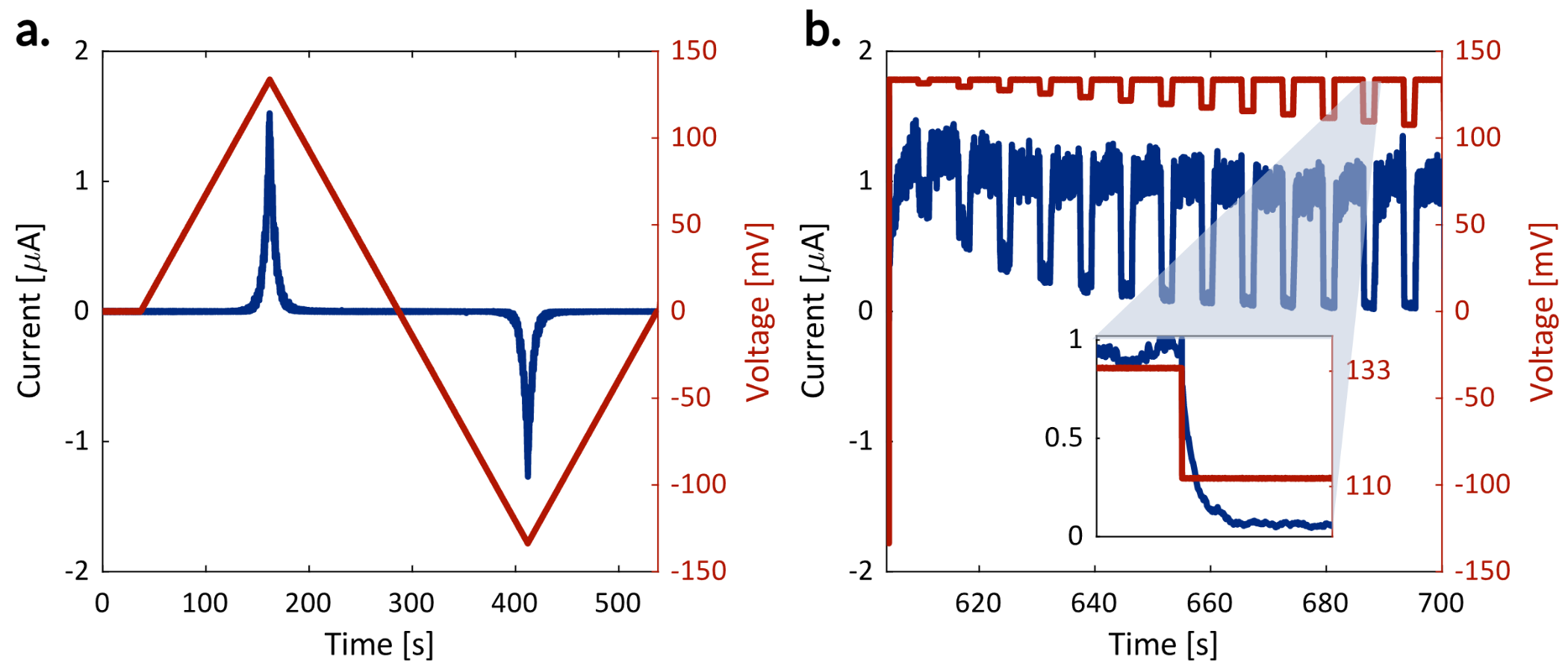}
\caption{Sample input waveforms and device responses used for the parameter fitting routine. a) The quasi-steady-state response of the device to a slow voltage sweep at 2 mHz. b) The voltage waveform and response were used to fit $\tau$ to a voltage function. The inset displays a close-up of the exponential decay of current in response to a step decrease in voltage. Exponential fits were computed for each voltage drop, and $\tau$ values corresponding to the voltage levels were recorded. Finally, these $\tau$ values were fitted to an exponential function of voltage, with the parameters denoted as $\tau_0$ and $V_\tau$.}
\end{figure}

\begin{figure} [H]
\includegraphics[width=\linewidth]{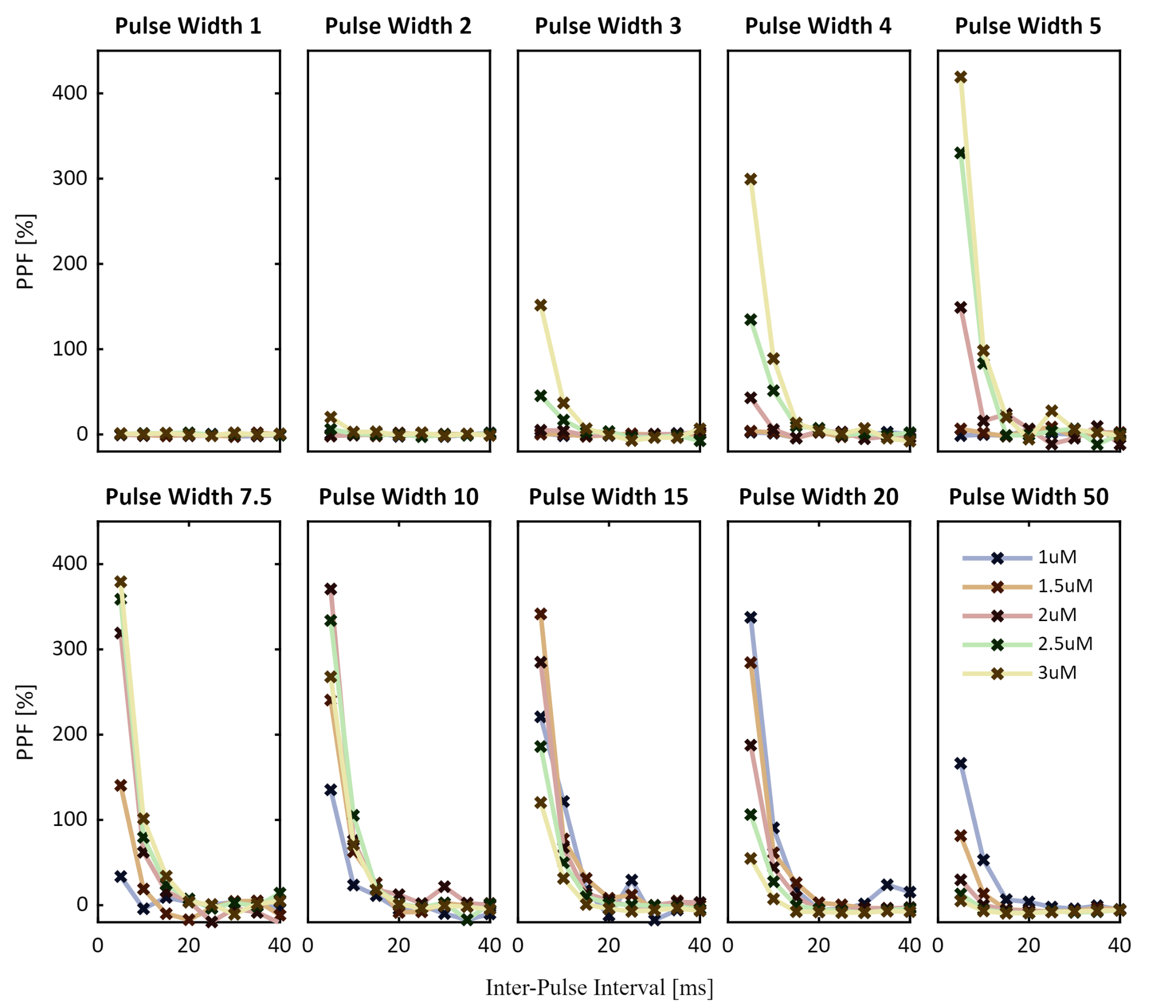}
\caption{Paired Pulse Facilitation (PPF) characteristics were examined for all five memristors as a function of pulse width and inter-pulse interval. For pulse widths below a certain threshold (which varies depending on the alamethicin concentration), no facilitation occurs because the memristors do not respond quickly enough. Conversely, there is little facilitation for very long pulse widths, as the system reaches a steady state for each pulse. Pulse widths falling between these extremes exhibit high PPF for specific memristors. Memristors with higher alamethicin concentrations respond more rapidly to relatively short pulses and peak their response at around a 5-ms pulse width. As the pulse width continues to increase, the lower concentrations display relatively higher PPF, with the lowest concentration showing the highest relative response.}
\end{figure}

\begin{figure} [H]
\includegraphics[width=\linewidth]{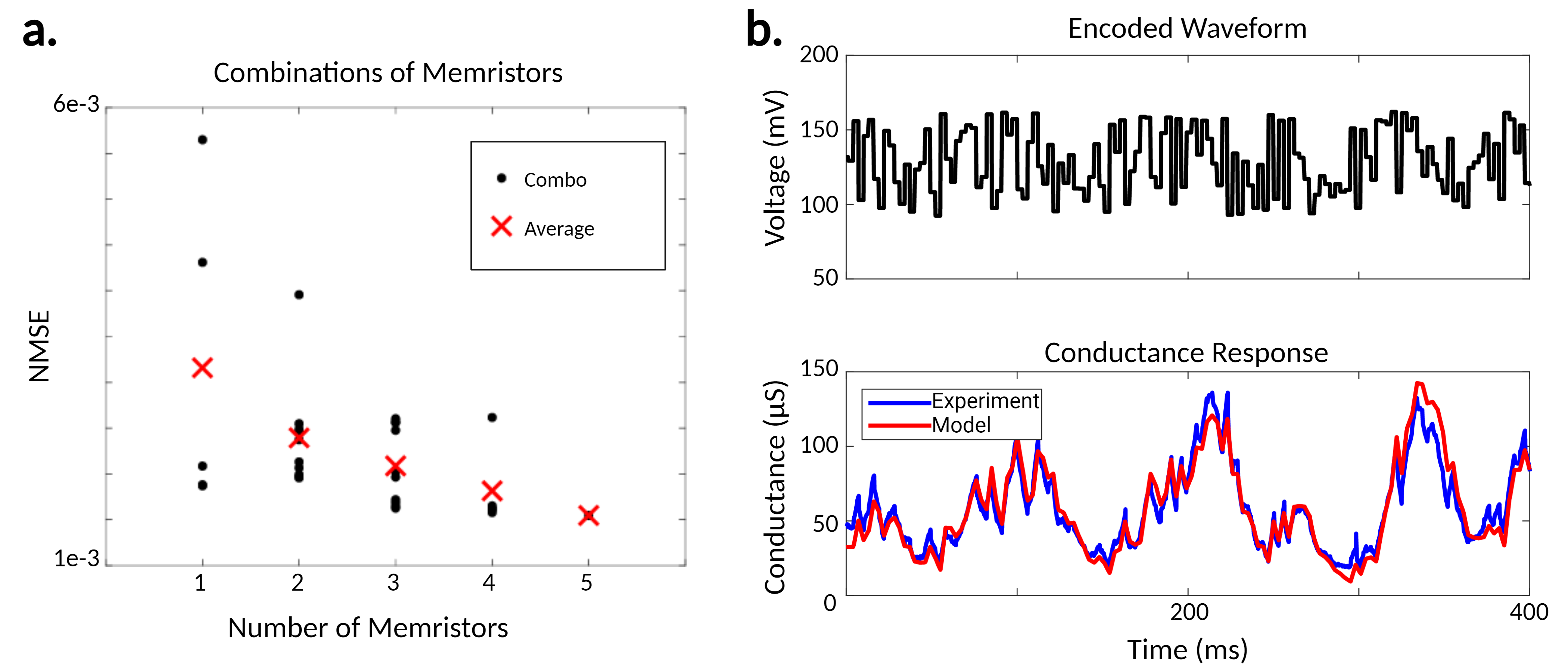}
\caption{(a) Experimental improvement of the NMSE as the number of varied memristors in the reservoir is increased from 1 to 5 devices. (b) A comparison of the conductance of a memristor in simulation vs. experiment (bottom) when subjected to the same voltage waveform (top).}
\end{figure}

\begin{figure} [H]
\includegraphics[width=\linewidth]{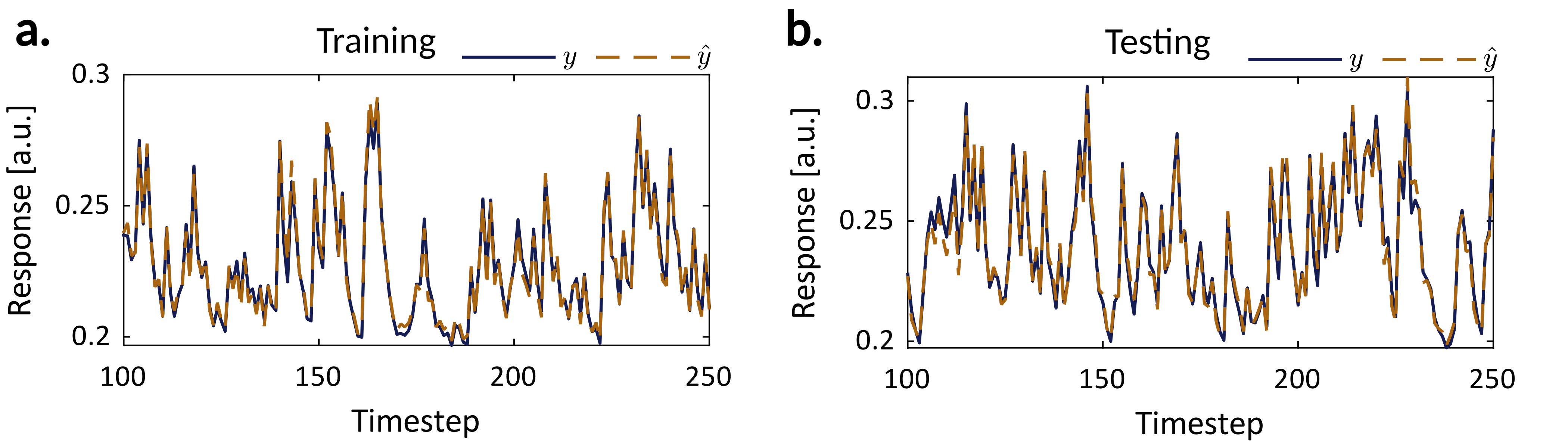}
\caption{The simulation results are based on a 5-memristor reservoir, with each memristor using corresponding parameters of 1 $\mu$M, 1.5 $\mu$M, 2 $\mu$M, 2.5 $\mu$M, and 3 $\mu$M alamethicin, respectively. (a) A comparison between the predicted $\hat{y}$ value of the reservoir during the training phase and the output of the SONDS equation, $y$. The training NMSE value was determined to be 1.64e-4. (b) Testing results are obtained using the weight vector derived from linear regression on the training data. The testing NMSE was calculated to be 2.18e-4.}
\end{figure}
\begin{figure} [H]

    \includegraphics{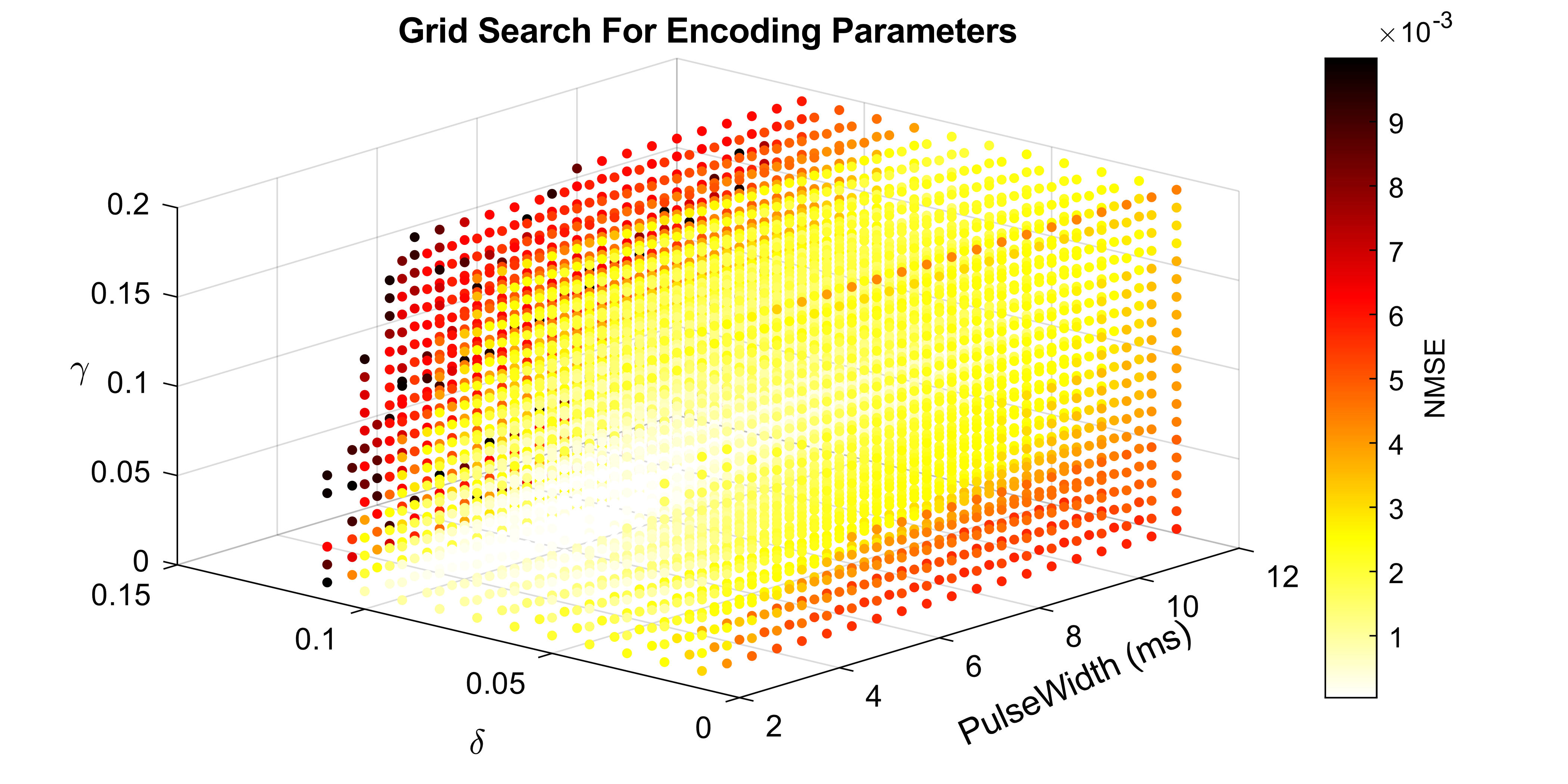}
    \caption{A heatmap visualization of the simulated 3-dimensional grid search for encoding parameters in the SONDS task. In this plot, $\gamma$ and $\delta$ are the scaling and offset parameters, respectively. The grid search shown here represents a 20x20x20 coarse grid over the parameter space. The results indicate a low pulse width of around 3 ms, an offset of 50 mV, and a scaling factor of 70 mV produces the best outcome in simulations. The scaling factor was increased to 100 mV in experiments to increase the signal-to-noise ratio.}
    
\end{figure}

\begin{figure} [H]
\includegraphics[width=\linewidth]{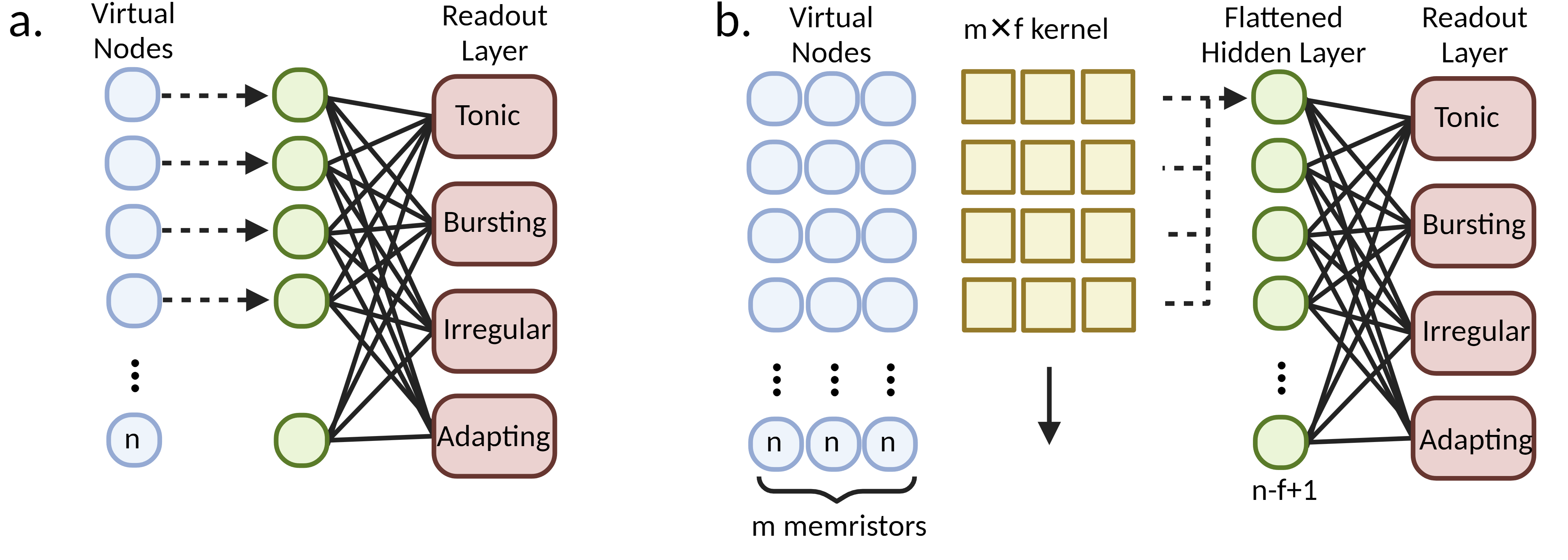}
\caption{ (a) A diagram describing the FC where the number of parameters is $4n+4$ to train the readout layer. (b) A diagram describing the 2-D convolution applied to create a hidden layer after the reservoir in the multi-memristor reservoir computing system. Here, $n$ is the number of virtual nodes, $m$ is the number of memristors that comprise the reservoir layer, and $f$ is the kernel length. Here, the number of parameters is $4(n-f+1)+4+m\times f+1$ which can substantially reduce the number of trained weights compared to the fully connected layer.}
\end{figure}

\begin{figure} [H]
\includegraphics[width=\linewidth]{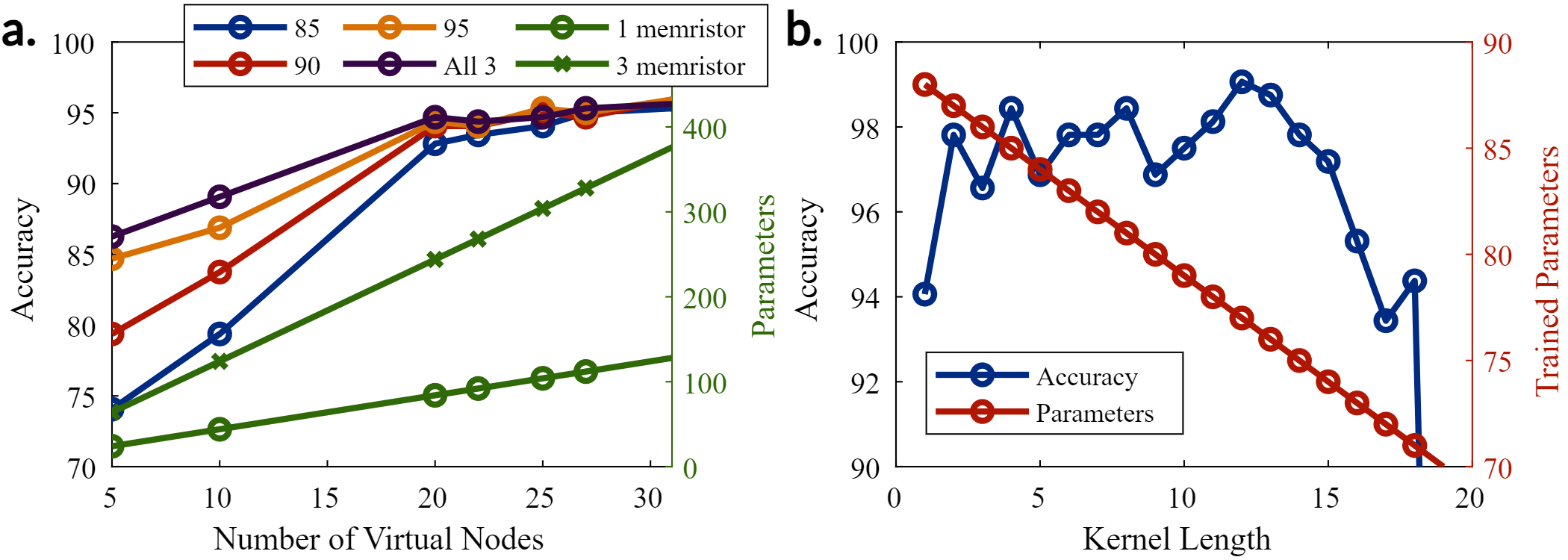}
\caption{(a) The graph shows the accuracy of a simulated reservoir with an FC readout layer for 85, 90, and 95 mV offsets and all three together and the number of trained parameters for each. Based on this plot, the accuracy begins to plateau at 20 virtual nodes. (b) The graph shows the accuracy and number of trained parameters of a simulated three-memristor reservoir with a 2-D convolution layer, where 20 virtual nodes were taken per memristor. This reservoir is made of all three offsets (85, 90, and 95 mV) and the kernel of the CNN is $3\times f$ where f is the kernel length, and it is varied to maximize accuracy and minimize the number of trained parameters.}
\end{figure}

\begin{figure} [H]
\includegraphics[width=\linewidth]{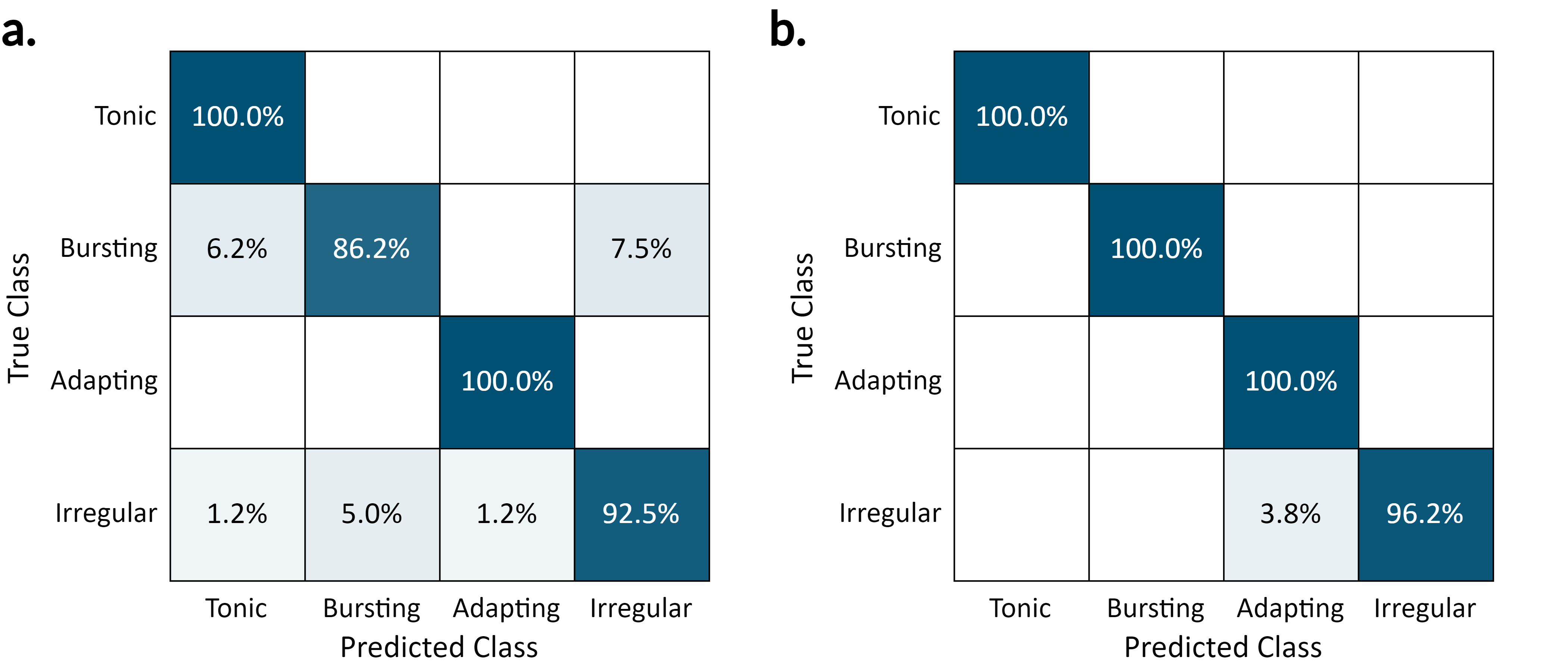}
\caption{ (a) Confusion matrix of simulated three memristor reservoir with FC readout layer. The overall accuracy of the best FC readout layer was $94.68\%$ and there were 240 trained parameters. (b) Confusion matrix of simulated three memristor reservoir with 2-D convolutional hidden layer. The overall accuracy of the best 2-D simulated hidden convolution layer was $99.06\%$ and there were 77 trained parameters.}
\end{figure}

\begin{figure} [H]
\includegraphics[width=\linewidth]{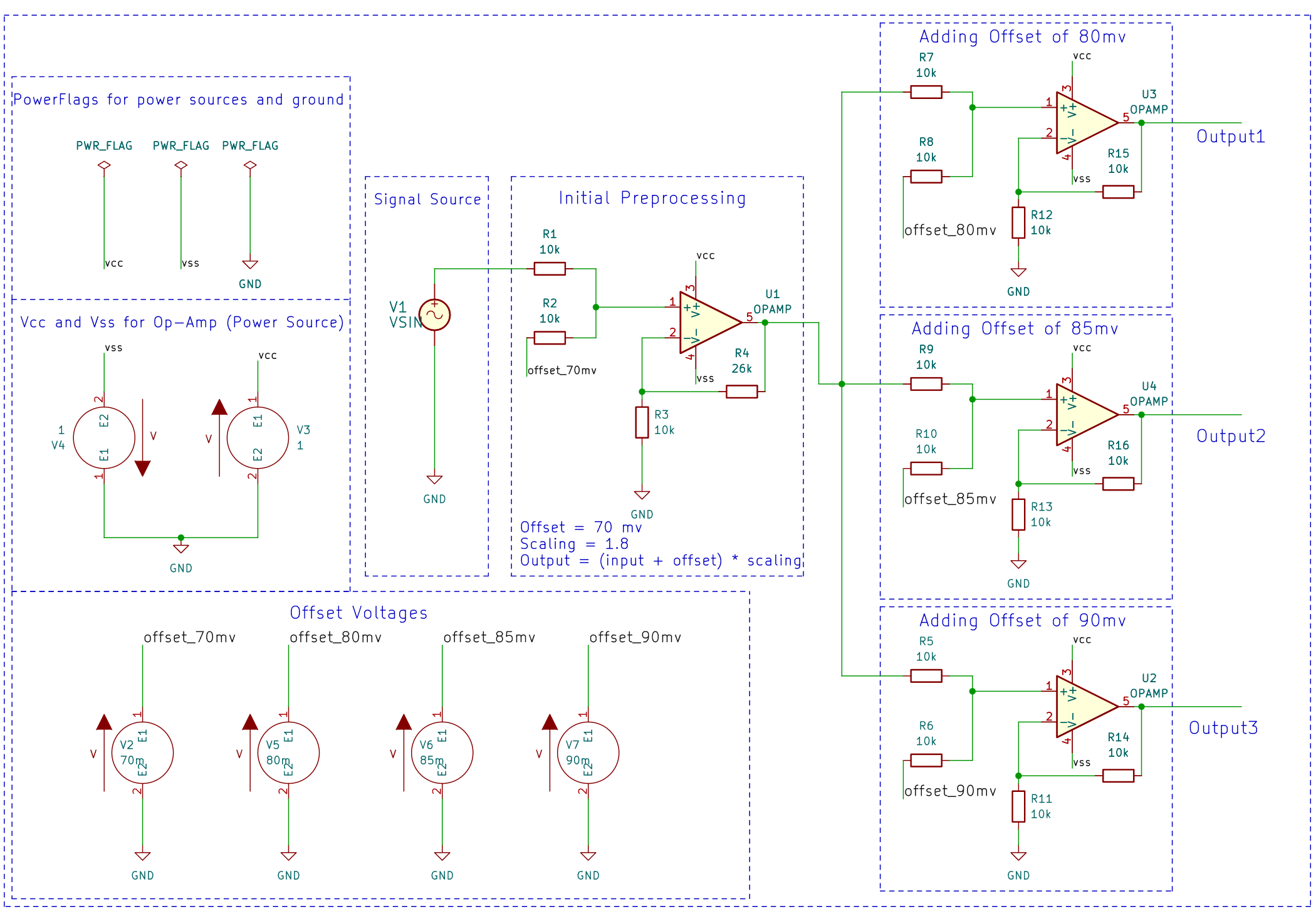}
\caption{ A Circuit diagram of the amplification and bias stage of neural activity analysis. This all-analog implementation converts biological action potentials in the range of -70 mV to +40 mV to a range of +70 mV to +280 mV. Note that this change in waveform is a linear scaling with DC offset so no temporal changes in the data occur.}
\end{figure}

\begin{figure} [H]
\includegraphics[width=\linewidth]{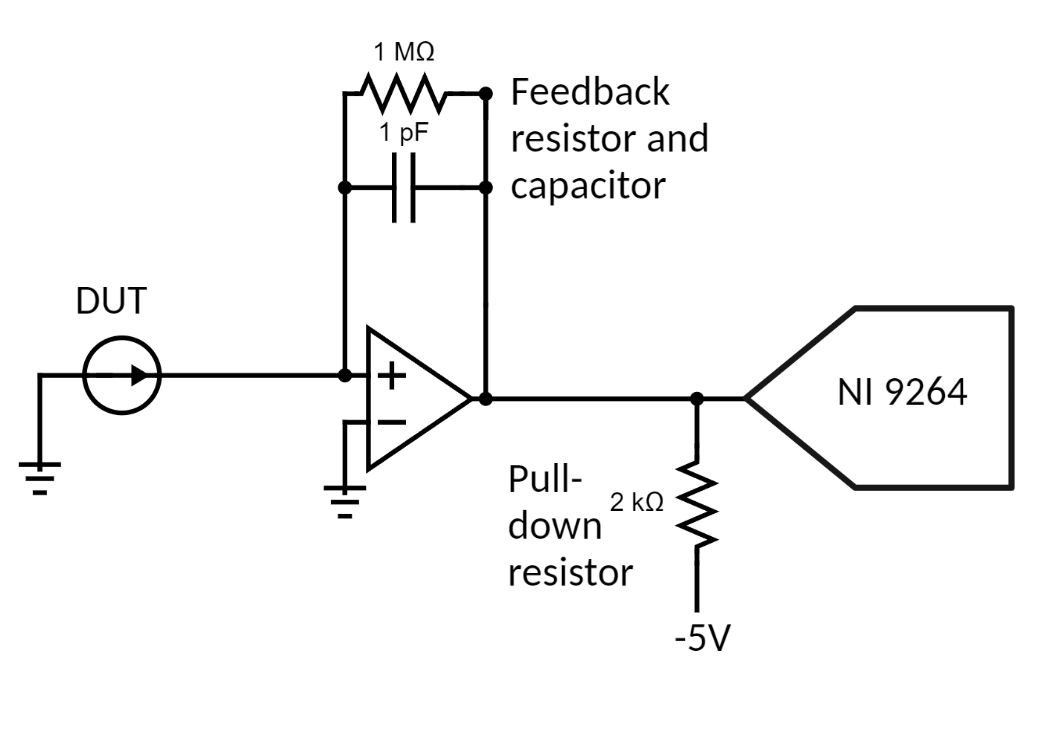}
\caption{ a Circuit diagram of the transimpedance amplifier (TIA) used to measure the current through memristors when the voltage exceeded the 200 nA maximum of the patch clamp amplifier. The op-amp used is Texas Instruments OPA380, which operates at 5V and draws 3 to 10 mA. This circuit performs current-to-voltage conversion as well as amplification. The gain shown is 1e6 and currents between 0 and 5 $\mu$A are converted to voltages between 0 and 5 V. The stabilizing feedback capacitor prevents oscillation of output. The pull-down resistor and a -5 V secondary supply allow the TIA response to swing to slightly below 0 V. The RMS voltage noise is 0.8 mV corresponding to 0.8 nA of current noise.}
\end{figure}

\begin{figure} [H]
\includegraphics[width=\linewidth]{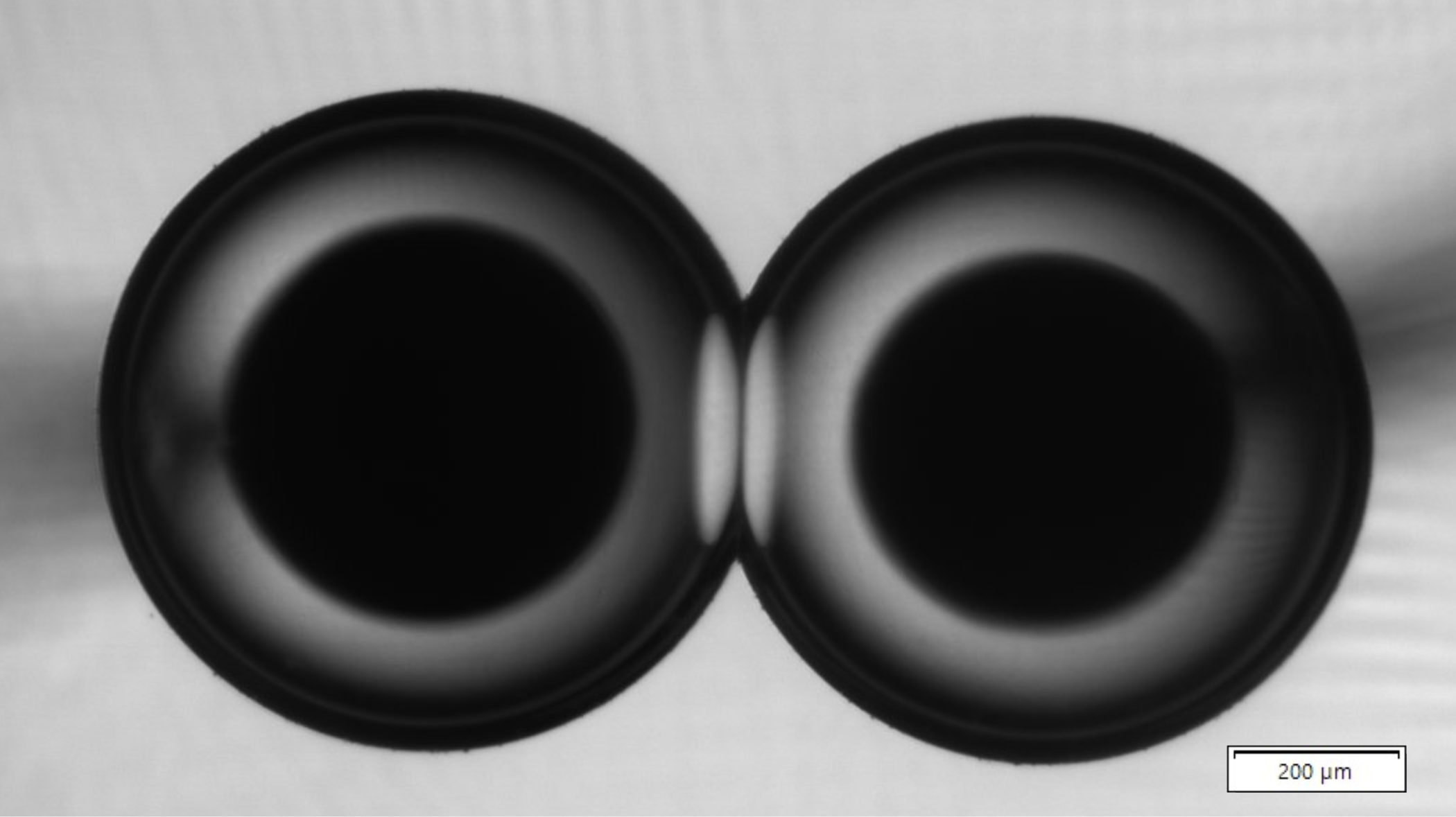}
\caption{ A bottom-view micrograph of the droplet interface bilayer setup using the hanging-drop method. The aqueous droplets are submerged in hexadecane oil and each contains 2mg/mL DPhPC, 1M KCl, and 10 mM MOPS. The electrodes we used were Ag/AgCl and electrical interrogation was performed with either an Axopatch 200b or the custom TIA shown in Figure S9.}
\end{figure}

\section{Supporting Discussion}

\subsection{Model Description}
The conductance of the alamethicin-based memristor is directly proportional to the number of pores formed by the alamethicin peptides within the insulating DPhPC membrane when exposed to a suprathreshold voltage. Furthermore, as voltage is applied, the membrane area increases due to electrowetting, leading to a further increase in the conductance. This phenomenon can be modeled using the following equation: 
\begin{equation}
    \label{2}
    I=G(N_a,A)V
\end{equation}
where $N_{a}$ is the number of alamethicin pores per unit area, and $A$ is the area of the lipid membrane. Furthermore, the conductance, $G(N_a,A)$, can be expressed as follows:
\begin{equation}
    \label{3}
    G=G_{u}N_{a}(V,t)A
\end{equation}
where $G_{u}$ is the average conductance per alamethicin pore\cite{najem2018memristive}. Based on the works of Najem et al. \cite{najem2018memristive}, the state equation for $N_{a}$ can be expressed as a first-order system:
\begin{equation}
\label{4}
\frac{dN_{a}}{dt}=\frac{1}{\tau}(N_{SS}-N_a)
\end{equation}
where $\tau$ is the time constant and $N_{SS}$ is the steady-state number of open pores per area\cite{eisenberg1973nature}. These two terms are also exponential functions of voltage and the full equation can be expanded as:
\begin{equation}
    \label{5}
    \frac{dN_{a}}{dt}=\frac{1}{\tau_{0}e^\frac{V}{V_{\tau}}}(N_{0}e^\frac{V}{V_{e}}-N_{a})
\end{equation}
where $\tau_{0}$, $N_{0}$, $V_\tau$, and $V_e$ are the time constant with $0 V$ applied, the number of pores open at $0 V$, the voltage required to drive an $e$-fold increase in $\tau$, and the voltage required to drive an $e$-fold increase in $N_{SS}$ respectively. These four parameters determine both the nonlinearity and the time constant of the memristor's conductance response. The model of an alamethicin-based memristor is described in more detail by Eisenberg et al. \cite{eisenberg1973nature} and Najem et al. \cite{najem2018memristive}. As an extension of the Eisenberg model, we introduce an additional condition into the model. This condition modifies certain parameters when the applied voltage surpasses the insertion threshold, $V_{T}$, as follows:
\begin{equation}
    \label{6}
    \frac{dN_{a}}{dt}=
        \begin{cases}
             \frac{1}{\tau_{01}e^\frac{V}{V_{\tau1}}}(N_{0}e^\frac{V}{V_{e}}-N_{a}) & \text{if } V < V_{T}\\
             \frac{1}{\tau_{02}e^\frac{V}{V_{\tau2}}}(N_{0}e^\frac{V}{V_{e}}-N_{a}) & \text{if } V>V_{T}\\
        \end{cases}
\end{equation}
where $V_{T}$ is the empirically-determined ion channel insertion threshold, and $\tau_{01}$, $\tau_{02}$, $V_{\tau1}$, and $V_{\tau2}$ are the parameters of the model below and above the threshold, respectively. These additional terms and the conditional statement allow the model to fit experimental data at relatively high voltages and frequencies. Boheim \textit{et al.} \cite{boheim1978analysis} described a similar implementation of the alamethicin model. In this study, we investigate how variations in material composition can impact these seven parameters, resulting in distinct temporal responses for each composition under identical inputs. Specifically, we manipulate the concentration of alamethicin peptides and perform parameter fitting for each concentration, as outlined in \textbf{Equation 5} (see Methods section).

\subsection{Model Fitting}
To find the parameters for exactly 1 M KCl and 10 mM MOPS, as well as various alamethicin concentrations, a consistent parameter-fitting routine was required. The fitting procedure, as described in the main text and elaborated here, was followed to generate the parameters presented in Table S1. In Figure S4, a sample of the waveforms used to stimulate the devices and an example of the recorded response is displayed. Initially, a slow voltage sweep at 2 mHz was chosen to determine the steady-state conductance relationship with voltage. The symmetric response observed in both the rising and falling voltage regions, as shown in Figure S2a, indicates that the sweep rate is slow enough to achieve an approximately steady-state conductance at each voltage point swept over. Sweeping at a faster rate, such as 200 mHz, results in an asymmetric conductance response, as seen in Figure S1. The slow sweep current response is divided by the applied voltage at each time step to calculate conductance values. These conductance values are then further divided by the area of the DIB and the conductance of an average alamethicin pore to determine the number of alamethicin pores per unit area. The fitting process involves applying a simple exponential function:

\begin{equation}
    N_{ss}(V)=N_{0}e^\frac{V}{V_{e}}
\end{equation}
This equation approximates the steady-state number of pores per area, and it is very accurate for large values of $N$. However, it is non-zero for all subthreshold voltages, which in reality, will have precisely zero pores formed. Therefore, in practice, this model is impractical for use with sub-macroscopic currents (~10 nA). Despite this, this model assumption faithfully reproduces large conductance responses in response to a wide range of voltages as shown in Figure S1. Next, the time constants were fit to a function of voltage as well. The time constants, as explained in the main text, are determined by reducing the voltage from an initially held higher value to a lower inspection voltage. Figure S2b shows a sample of the voltage waveform and current response. These current responses were divided by the voltage applied to obtain conductance. The time constant of the decay was found by fitting the response to a simple exponential function in time:
\begin{equation}
    G(t)=ae^\frac{-t}{\tau}
\end{equation}
where $G$, $t$, $a$, and $\tau$ are the conductance, time, a fitted scaling parameter, and the time constant, respectively. In the original formulations of the alamethicin model, the time constants had a similar relationship to voltage as the steady-state number of inserted pores:
\begin{equation}
    \tau(V)=\tau_{0}e^\frac{V}{V_{\tau}}
\end{equation}
For low voltages and low concentrations, this simple function is sufficient to explain the empirical relationship\cite{najem2018memristive,okazaki2003ion,eisenberg1973nature}. Nevertheless, we noticed a deviation from this relationship at elevated voltages and conductances, which became more pronounced at even higher voltages. As such, we split the fitting into two separate conditional sections. From applied voltages below the threshold, $V_{T}$, we fit a single exponential relationship as follows:
\begin{equation}
    \tau(V)=\tau_{01}e^\frac{V}{V_{\tau1}}
\end{equation}
On the other hand, for voltages above the switching threshold, $V_{T}$, we fit a separate exponential as follows:
\begin{equation}
    \tau(V)=\tau_{02}e^\frac{V}{V_{\tau2}}
\end{equation}
This adjustment allowed our model to successfully predict the memristors' conductance response to long waveforms of medium to high voltages (Figure S4b) by increasing model fidelity in a larger region of input values.
Finally, due to constructing lipid bilayers using the hanging-drop method, bilayer area changes as a function of voltage \cite{najem2018memristive}. In operations at near-constant voltages (varying by only a few tens of mV), the change in bilayer area is negligible. However, when fitting for the model parameters, wide ranges of voltages are applied, and the change in membrane area is significant. This is relevant because the conductivity of the membrane scales linearly with increased area, and these bilayer area changes are also transient in time. To account for these changes and to exclude their effect on the model, a video recording of the droplet interface bilayers was taken while the fitting waveforms were applied. Changes in area were determined by visual changes in the contact diameter between the two droplets. Videos synchronized with the voltage waveforms allowed us to divide the measured conductance at all time steps by the membrane area to find the conductance per area, $G_a$. Furthermore, we divided $G_a$ by $G_u$, which represents the average conductance per alamethicin pore, with the specific value of $G_u$ being sourced from the literature \cite{okazaki2003ion}. The resulting normalized value, $N_a$, represents the number of pores per unit area and is the variable evaluated in the state equation (Supplemental Equation 5)

\subsection{SONDS task and signal-to-noise ratio}
As noted in the main text, the noiseless simulation of 5 memristors could form a reservoir that was significantly more accurate in predicting the second-order task than the experimental results. Figure S5 shows the training and testing data for the simulated memristor reservoir. The increased accuracy can, in part, be attributed to low-alamethicin-concentration memristors, which operate at significantly lower conductances compared to high-concentration memristors. The measurement noise remained consistent across all memristors since they were assessed using the same hardware. However, due to the lower absolute conductance in low-concentration memristors, their signal-to-noise ratio was inherently diminished, resulting in a less reliable contribution to the reservoir layer. This effect was observed in single memristor reservoirs. The highest NMSEs were recorded for high alamethicin concentrations. Nonetheless, simulations indicate that when all memristors exhibit equally strong signal-to-noise ratios, the advantages of a more diverse reservoir become more evident. Furthermore, this signal-to-noise limitation prevented us from utilizing the best-predicted encoding parameters in the experimental implementation (as shown in the bright white region of Figure S6). The parameters used in the experimental section were optimized considering this practical constraint. However, the existence of even better simulation results suggests further potential for optimizing memristor reservoirs.

\section{Supporting Table}
\begin{table} [H]
 \caption{Empirically-found parameters for each concentration of alamethicin. Values from this table are used in Equation 6 to model the change in the number of pores per area of the memristors.}
  \begin{tabular}[htbp]{@{}llllllll@{}}
    \hline
    [Alm]  & $V_e$ & $N_0$  & $V_{\tau 1}$ & $\tau_{01}$ & $V_{\tau 2}$  & $\tau_{02}$ & $V_{T}$\\
    ($\mu$M) & (mV) & ($pores.m^{-2}$) & (mV) & (ms) & (mV) & (ms) & (mV)\\
    \hline
    1.0 & 5.4 & 0.044 & 45.4 & 1 & 11.4 & 0.00085 & 107\\
    1.5 & 5.5 & 1.3 & 45.4 & 1 & 14.2 & 0.017 & 85\\
    2.0 & 5.6 & 5.4 & 46.4 & 1.1 & 13.9 & 0.019 & 79\\
    2.5 & 5.5 & 22.4 & 44.4 & 1 & 16.5 & 0.076 & 69\\
    3.0 & 5.7 & 140 & 43.2 & 1.1 & 19.0 & 0.2 & 57\\
    \hline
  \end{tabular}
\end{table}

\begin{table} [H]
 \caption{Results for various combinations of memristors in the reservoir layer that are optimized in the CNN layer by changing the kernel size to produce the highest accuracy.}
  \begin{tabular}[htbp]{@{}llll@{}}
    \hline
    Offsets (mV) & Kernel Size & Accuracy ($\%$) & Parameters\\
    \hline
    85 & $1\times 13$ & 81.5 & 50\\
    90 & $1\times 11$ & 93.1 & 56\\
    95 & $1\times 13$ & 93.7 & 50\\
    85,90 & $2\times 2$ & 95.3 & 85\\
    90,95 & $2\times 6$ & 95.0 & 77\\
    85,95 & $2\times 8$ & 95.3 & 73\\
    85,90,95 & $3\times 9$ & 96.5 & 80\\
    \hline
  \end{tabular}
\end{table}

\bibliography{Library2}